\documentclass[english,journal,ruled,vlined]{article}
\usepackage[T1]{fontenc}
\usepackage[latin9]{inputenc}
\usepackage{geometry}
\usepackage{verbatim}
\usepackage{color}
\usepackage{babel}
\usepackage{algorithm2e}
\usepackage{amsmath}
\usepackage{amsthm}
\usepackage{amssymb}
\usepackage{graphicx}
\usepackage{setspace}
\usepackage{nomencl}
\usepackage{float}

\usepackage[unicode=true,
 bookmarks=false,
 breaklinks=false,pdfborder={0 0 1},backref=section,colorlinks=false]
 {hyperref}

\makeatletter

\providecommand{\tabularnewline}{\\}

\theoremstyle{plain}
\newtheorem{thm}{\protect\theoremname}
\theoremstyle{plain}
\newtheorem{prop}[thm]{\protect\propositionname}

\usepackage{babel}

\newcommand{\splitatcommas}[1]{%
  \begingroup
  \ifnum\mathcode`,="8000
  \else
    \begingroup\lccode`~=`, \lowercase{\endgroup
      \edef~{\mathchar\the\mathcode`, \penalty0 \noexpand\hspace{0pt plus 1em}}%
    }\mathcode`,="8000
  \fi
  #1%
  \endgroup
}

\usepackage{soul}
\date{}

\usepackage{multirow}

\usepackage{babel}

\def\nomname{Nomenclature}

\@ifundefined{showcaptionsetup}{}{%
	\PassOptionsToPackage{caption=false}{subfig}}
\usepackage{subfig}
\makeatother

\def\nomname{Nomenclature}

\providecommand{\propositionname}{Proposition}
\providecommand{\theoremname}{Theorem}
\providecommand{\propositionname}{Proposition}
\providecommand{\theoremname}{Theorem}

\begin{document}
\title{Deep Spatio-temporal Sparse Decomposition for Trend Prediction and
Anomaly Detection in Cardiac Electrical Conduction}
\author{Xinyu Zhao$^{1}$, Hao Yan$^{1}$, Zhiyong Hu$^{2}$, Dongping Du$^{2}$
\thanks{This work was supported in part by National Science Foundation under
Grant DMS-1830363, CMMI-1646664, CMMI 1922739, and CMMI-1728338.} \thanks{Xinyu Zhao and Hao Yan $^{1}$ are with School of Computing, Informatics,
and Decision Systems Engineering, Arizona State University, Tempe,
AZ 85281, USA. (e-mail: haoyan@asu.edu). } \thanks{Zhiyong Hu and Dongping Du $^{2}$ are with Industrial, Manufacturing
\& Systems Engineering in Texas Tech University, Lubbock, TX 79409,
USA (email: Dongping.Du@ttu.edu)} }
\maketitle
\begin{abstract}
Electrical conduction among cardiac tissue is commonly modeled with
partial differential equations, i.e., reaction-diffusion equation,
where the reaction term describes cellular stimulation and diffusion
term describes electrical propagation. Detecting and identifying
of cardiac cells that produce abnormal electrical impulses in such
nonlinear dynamic systems are important for efficient treatment and
planning. To model the nonlinear dynamics, simulation has been widely
used in both cardiac research and clinical study to investigate cardiac
disease mechanisms and develop new treatment designs. However, existing
cardiac models have a great level of complexity, and the simulation
is often time-consuming. We propose a deep spatio-temporal sparse
decomposition (DSTSD) approach to bypass the time-consuming cardiac
partial differential equations with the deep spatio-temporal model
and detect the time and location of the anomaly (i.e., malfunctioning
cardiac cells). This approach is validated from the data set generated
from the Courtemanche-Ramirez-Nattel (CRN) model, which is widely
used to model the propagation of the transmembrane potential across
the cross neuron membrane. The proposed DSTSD achieved the best
accuracy in terms of spatio-temporal mean trend prediction and anomaly
detection. 
\end{abstract}

\section{Introduction \label{Sec: Intro}}

Cardiac arrhythmia is a group of conditions where the heartbeat is
irregular. Common conditions include ventricular tachycardia, atrial
flutter, ventricular fibrillation, etc. These conditions occur when
abnormal and chaotic electrical impulses cause heart chambers to quiver
ineffectively instead of pumping blood to support the body. Such abnormal
activities can result in serious complications such as stroke and
even sudden death. For arrhythmias that can not be treated with medications,
surgical procedures can be done to locate abnormal cardiac cells that
initiate disorders and burn the tissue to stop the abnormal electrical
activities. However, the identification of abnormal electrical impulses
and problematic tissue is challenging.

Detection and identification of the cardiac cells that cause arrhythmia
can be defined as an anomaly detection problem. Anomaly detection
is not a new concept in cardiac research. Many studies have been done
to identify dissimilar heartbeats in Electrocardiogram (ECG) to aid
cardiac diagnosis. The majority of these studies use signal processing
and machine learning techniques to distinguish unusual waveforms in
ECG and detect abnormal cardiac events \cite{kropf2018cardiac}. In
addition, works have been done to detect anomalies in time series
of multi-parameter clinical data to distinguish critical from non-critical
conditions for patients undergoing heart surgery \cite{presbitero2017anomaly}.
Most of the existing studies focus on analyzing raw time series data
to detect abnormal patterns. Currently, the identification of cardiac
cells that initiate and maintain irregular electrical activities remains
challenging. This paper focuses on detecting and locating dissimilar
cellular stimulation (i.e., \textit{anomaly}) from a large number
of \textit{normal} cells whose transmembrane potentials either stay
at a constant value or vary regularly in a normal way by analyzing
signals (i.e., changes of transmembrane potential overtime) generated
from individual cells. Detecting when and which cells initiate the
abnormal electrical impulses is important for efficient treatment
design and planning. More literature on anomaly detection based on
the cardiac electrical conduction will be discussed in Section \ref{subsec: CardiacModel}.

There are significant challenges involved in detecting the anomaly
from normal cardiac electrical activities. The first challenge is
that normal cardiac activities usually present very \textit{complicated
spatio-temporal patterns}. To study such patterns, computer models
across different organizational scales, including cellular models,
tissue models, and organ models, are often used. The cardiac cell model
describes the transmembrane potential as a function of time by an
ordinary differential equation, and the measurement of the transmembrane
potential waveform signifying the electrical activity of cardiac cells
is known as action potential \cite{xie2002electrical}. The propagation
of electrical waves in cardiac tissue is modeled by a reaction-diffusion
equation where the reaction term describes the cellular stimulation,
and the diffusion term represents cell-to-cell interactions. For example,
the Courtemanche-Ramirez-Nattel (CRN) model was developed \cite{courtemanche1998ionic},
which consists of over seventy coupled equations to describe cardiac
depolarization and repolarization.

The second challenge is that anomaly can happen at any location, time,
and magnitude with complex spatio-temporal propagation behavior. This
paper aims to model the premature firing of the transmembrane potentials
due to either abnormal diastolic depolarization or after repolarization,
which is one of the major causes of cardiac arrhythmia, such as atrial
fibrillation. In the experiments, the regular/periodic stimulation
triggers the normal electrical conduction, and the irregular/random
stimulation is given to trigger abnormal stimulation. We are interested
in detecting abnormal stimulation from regular and periodic
electrical activities. Anomaly, due to the remodeling of individual
cardiac cells, is often sparse. One specific challenge is that if
the anomaly is not detected timely after its initiation, it will be
propagated throughout the entire system and become hard to identify
the exact time and location when or where it starts.

The third challenge is that although many simulation models such as
CRN provides accurate quantification of cardiac cellular functions,
they may suffer from high computational time and unknown parameters.
For example, when the cardiac model extends to the higher organizational
scales, e.g., tissue and organ scales, the simulation is very time-demanding.
Furthermore, the identification and customization of the CRN model
to specific applications are challenging since it involves unobservable
variables (e.g., ion channel gating variables), which are difficult
to measure in in-vitro/in-vivo experiments.

To address these challenges, we propose to learn a metamodel to replace
the time-consuming simulation models. Cardiac electrical propagation
is inherently a spatio-temporal process with transmembrane potential
changing in a nonlinear fashion in the temporal domain and electrical
waves propagating in the spatial domain. Therefore, deep spatio-temporal
models can be constructed to learn the hidden dynamics in the spatio-temporal
processes. Deep neural networks such as convolutional neural networks
\cite{RN7} and recurrent neural networks \cite{RN8} have been proved
as efficient models to describe the complex spatio-temporal processes.
For example, deep learning has recently been used as a metamodel to
replace the traditional partial differential equations (PDE) in cardiac simulation and has achieved
great prediction accuracy \cite{yan2019physics}. This can be realized
by learning the hidden dynamics of the spatio-temporal process from
simulation data generated by realistic cardiac models.

In many cases, it is important not only to model the normal spatio-temporal
patterns but also to detect when and where the anomaly would happen
(i.e., the cardiac cells that produce irregular electrical impulses).
In literature, spatial-temporal smooth-sparse decomposition was proposed
to detect sparse anomalies from the smooth spatial and temporal mean
trend \cite{yan2018real}. However, due to the assumption of the linear
basis representation for the complicated spatio-temporal foreground,
it is not suitable to model complicated spatio-temporal patterns in
the cardiac electrical conduction. Furthermore, since the time interval
of such impulse is short and the anomaly pattern follows the same
spatiotemporal propagation rule (e.g., CRN equation) according to
the normal patterns. Therefore, it creates significant challenges
in anomaly detection in such complicated systems. How to apply such
deep learning methods for real-time spatio-temporal metamodeling and
anomaly detection has not been fully discussed yet in literature,
especially for cardiac electrical simulation. More discussions will
be provided in Section \ref{Sec: Literature}.

In this paper, we will focus on developing a new deep
spatio-temporal sparse decomposition (DSTSD) method, which combines
the power of the deep neural network to represent the complicated
spatio-temporal patterns of the mean trend and the decomposition framework
to separate the sparse anomaly from the mean trend. More specifically,
two spatio-temporal model architectures are combined into the proposed
DSTSD, namely the ConvLSTM and ConvWaveNet, for metamodeling and
model the nonlinear dynamics of the cardiac electric conduction. We
demonstrate that in the case of highly nonlinear spatio-temporal systems,
the proposed DSTSD method can achieve both the smallest detection
delay and accurate localization of the anomaly. 

In conclusion, the rest of the paper is organized as follows. Section
\ref{Sec: Literature} reviews the related literature in cardiac
electrical conduction modeling and spatio-temporal anomaly detection.
Section \ref{Sec: Motivating Study} gives the motivating example
of our study for transmembrane potential simulation. Section~\ref{Sec: methodology}
introduces the proposed DSTSD methodology for spatio-temporal metamodeling,
spatio-temporal mean trend prediction, and anomaly detection. Section~\ref{sec: Simulation Study} shows the simulation study to demonstrate
the performance of the proposed method for both long-term prediction
and anomaly detection. Section~\ref{sec: conclusion} concludes the
paper with future work.

\section{Literature Review \label{Sec: Literature}}

In this section, we will first review the literature on modeling and
anomaly detection of Cardiac Electrical Conduction. We will then
review some data-driven methodology on the monitoring and diagnosis
of spatio-temporal data.

\subsection{Cardiac Electrical Conduction Modeling and Anomaly Detection \label{subsec: CardiacModel}}

Modeling and analysis of irregular cardiac electrical conduction have
been widely studied in the literature, including the efficient numerical
simulation model \cite{collet2017temperature,kaboudian2019real}.
To detect anomalies in such complex spatio-temporal systems, we briefly
classified the current methodology applied to the cardiac electrical
conduction Modeling into model-based, metamodel-based and statistical-based
methodology.

In the first category, most of the works have been focused on using
a model-based control method to control the anomaly \cite{dubljevic2008studies}.
For example, Garzon et al. \cite{garzon2014continuous} proposed a
model-based continuous-time feedback control methodology to suppress
the anomaly. Marcotte and Grigoriev \cite{marcotte2016adjoint} proposed
an adjoint eigenfunction method to provide localization for the dynamics
and control of the unstable spiral wave. Some efforts to reduce the
computational complexity, including Galerkin projection \cite{garzon2014continuous}
and numerical approximation \cite{kaboudian2019real} have been proposed.
However, the major limitations of the model-based methodology are
that these models require all the complex dynamic models and parameters
to be known, which may not be feasible in practice.

In the second category, metamodeling has been a popular approach that
helps reduce model complexity with unknown dynamics models and overcome
computational challenges. Gaussian Process (GP) model has been
a popular choice to extract information from high-dimensional data.
Especially, the GP model is widely used to model the shape of the
action potential \cite{mirams2016uncertainty,chang2015bayesian,johnstone2016uncertainty}.
However, the major limitation of GP models is that GP models lack
the ability to perform long-term prediction and are often computationally
inefficient, which is not a good candidate for metamodeling and real-time
anomaly detection purpose.

In the third category, machine learning methods are applied to detect
irregular behavior. For example, Yang et al. \cite{Yang_2013} proposed
a classification model by combining a feature embedding technique
and a self-organizing map to classify different types of myocardial
infarction. For unknown anomaly detection, Loppini et al. \cite{loppini2019spatiotemporal}
proposed to use the statistical correlation functions to detect irregular
behaviors. Greisas et al. \cite{greisas2014detection} proposed to
Principal Component Analysis (PCA) for the detection of abnormal cardiac
activity. However, without a good metamodel, it is often hard to accurately
infer the time, location, and magnitude of the external stimulation.

\subsection{Spatio-temporal Anomaly Detection Literature \label{subsec: stmodel}}

Here, we will review some data-driven methodology on the monitoring
and diagnosis of spatio-temporal data. Current research in this area
can be classified into three groups: principal component analysis-based
approach, functional data analysis-based techniques, and deep learning-based
methods.

In the first group, principal component analysis (PCA) is one of the
most popular methods for spatio-temporal data dimension reduction
because of its simplicity, scalability, and data compression capability.
For example, PCA \cite{liu1995control}, multivariate functional PCA
\cite{paynabar2016change}, tensor-based decomposition method \cite{yan2015image},
multi-resolution PCA \cite{bakshi1998multiscale}, subspace learning
\cite{zhang2018dynamic} have been proposed to reduce the dimensionality
and then apply the control chart on the low-dimensional embedding
and the residual space. The main drawback of current PCA-based methods
is that they cannot be directly used for spatio-temporal data streams
with a time-varying mean.

The second category attempts to model the spatio-temporal data as functional
data by modeling the data structure by a set of known spatial
or temporal basis, kernel, and covariance structure. For example,
non-parametric methods based on local kernel regression \cite{zou2008monitoring,qiu2010nonparametric,zou2009nonparametric},
splines \cite{chang2010statistical} and wavelets \cite{paynabar2011characterization}
are proposed. Other works such as longitudinal data analysis \cite{Qiu2014,xiang2013nonparametric},
Gaussian process \cite{cheng2015video} are also proposed. However,
these methods do not directly model the structure of the anomaly.
Therefore, decomposition-based approaches have become popular due
to the ability to decompose the anomaly signals directly from the
complex spatio-temporal trend \cite{yan2018real}. However, one major
drawback of the decomposition method is that these methods assume
the spatio-temporal data can be represented by a set of known basis
or kernels, which failed to model the complicated spatio-temporal
structure of the signal.

To monitoring complicated spatio-temporal systems, deep learning methods
such as convolutional neural networks and generative adversarial networks have been applied. We can divide the current literature for
deep learning in spatio-temporal anomaly detection into two classes:
Unsupervised autoencoder approaches and supervised spatio-temporal
regression. For the autoencoder approaches, spatio-temporal autoencoders
\cite{zhao2017spatio} and Generative Adversarial Nets \cite{ravanbakhsh2017abnormal}
have been proposed to detect anomalous events. The autoencoder
approaches can dramatically reduce the dimensionality of the original
problems, and the monitoring statistic is often defined as a function
of the model residual. In literature, prediction-based deep learning
methods have also been used for anomaly detection. For example, feed-forward
convolutional networks are proposed in \cite{mathieu2015deep} for
video prediction by minimizing the mean square error of future prediction.
Another deep learning framework for anomaly detection \cite{munawar2017spatio}
utilizes the unsupervised learning method to extract features and
then detect irregularities through a prediction system. For either
supervised and unsupervised anomaly detection problems in literature,
no existing works exist to separate the anomaly signals from the original
spatio-temporal mean trend.

\section{Motivating Study: Electrical Propagation Through One-dimensional
Cell String based on Courtemanche-Ramirez-Nattel (CRN) Model \label{Sec: Motivating Study}}

In this study, Courtemanche-Ramirez-Nattel (CRN) model
\cite{courtemanche1998ionic} was used to simulate the transmembrane
potential of individual cardiac cells. CRN model is a detailed model
that describes the complex mechanism of cardiac electrical signaling
in human atrial cells. The mono-domain tissue model is adopted to
simulate the electrical wave propagation on a one-dimensional (1D)
cell string. CRN model is a physiologically realistic model for human
atrial cells, which provides a detailed description of ionic channel
gating. More specifically, it models the complex spatio-temporal dynamics
of transmembrane potential, which is defined as the difference in the
electric potential between the interior and the exterior of the biological
cell, by partial differential equations (PDE) defined in (\ref{eq: CRN_mu}).
\begin{align}
\frac{\partial u(t,s)}{\partial t} & =-\frac{I_{ion}\big(u(t,s),\mathbf{v}(t,s)\big)}{C_{m}}+D\frac{\partial^{2}u(t,s)}{\partial s^{2}}+c(t,s)\label{eq: CRN_mu}\\
\frac{\partial v(t,s)}{\partial t} & =h\big(u(t,s),\mathbf{v}(t,s)\big).\label{eq: unobservable}
\end{align}
Here, $t$ represents time and $s$ indicates spatial location. $u(t,s)$
is the transmembrane potential, $\mathbf{v}(t,s)$ is a vector of
variables associated with the ion channel conductance (gating variables).
$c(t,s)$ is the external stimulus, $C_{m}$ is the total capacitance,
and $D$ is the isotropic diffusion coefficient determined by gap
junction resistance, surface-to-volume ratio, and membrane capacitance
\cite{courtemanche1998ionic}. $I_{ion}$ is the summation of 12 different
ion channel currents, which are controlled by $u(t,s)$ and $\mathbf{v}(t,s)$
with over 70 equations. These equations are compactly represented
by $h\big(u(t,s),\mathbf{v}(t,s)\big)$ in \eqref{eq: unobservable},
and their detailed expressions as well as model parameters used for
the data generation can all be found in \cite{xie2002electrical}.
In reality, the transmembrane potential $u(t,s)$ can often be measured
(i.e., observable), but the ion currents $I_{ion}$ and the associate
hidden variable $\mathbf{v}(t,s)$ are often hard to obtain (i.e.,
unobservable). Although $I_{ion}$ and $\mathbf{v}(t,s)$ are unobservable,
it is known that they take into effect on the transmembrane potential
within different periods, such as the sodium currents affect the rising
of $u(t,s)$ while potassium currents influence its restoration. It
is such a phenomenon that motivates the proposed modeling structure
in the next section. The CRN model has been popularly used in many studies to simulate both
normal heart functions and cardiac disorders such as atrial fibrillation
\cite{harrild2000computer,xie2002electrical,zahid2016patient}. However,
due to its computational complexity, metamodeling techniques are important
to reduce computational complexity. The details of the CRN model can
be found in \cite{courtemanche1998ionic}.

\begin{figure}[h!]
\centering \subfloat[Original Spatio-temporal Map for Case 1]{\includegraphics[width=0.35\linewidth]{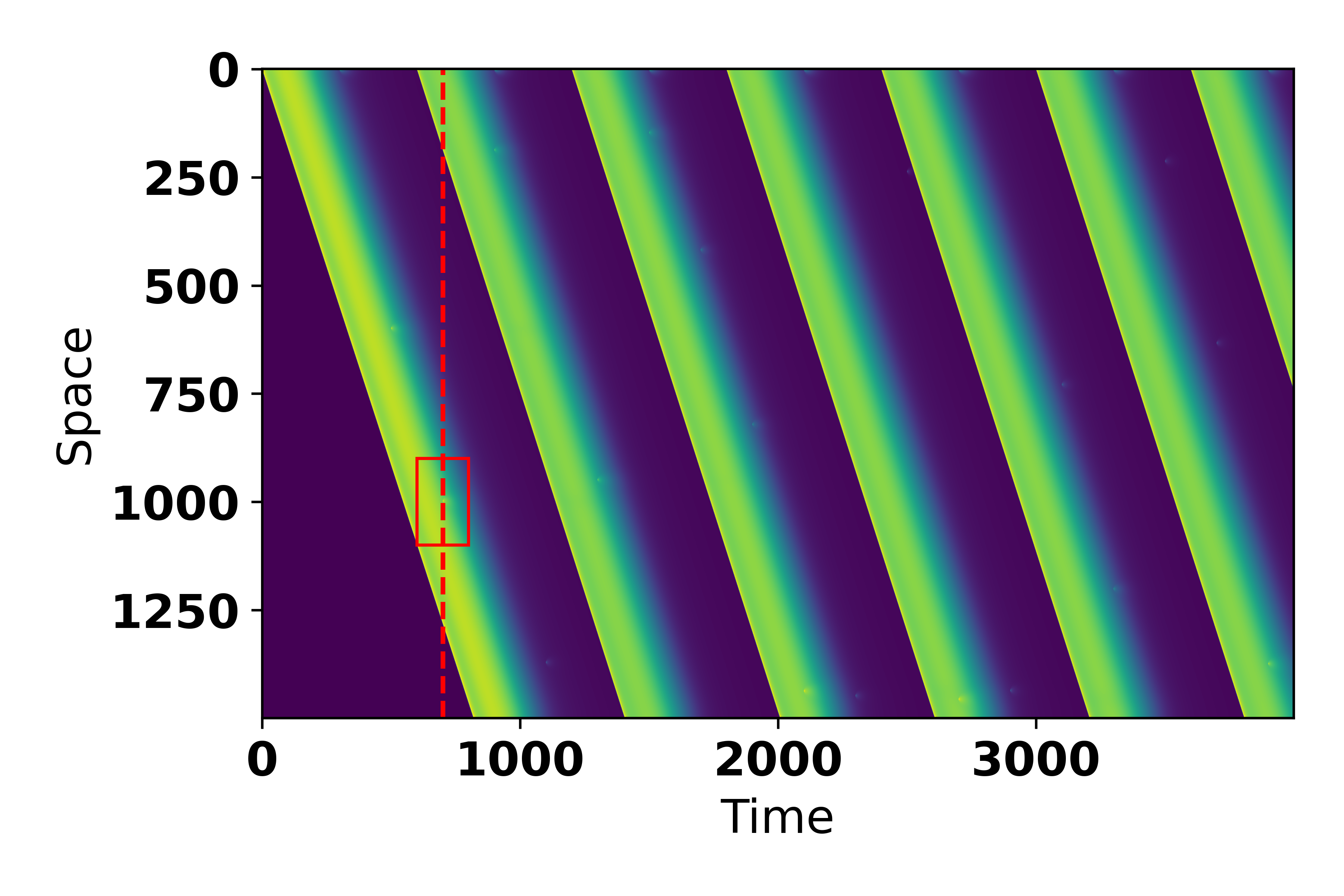} \label{case_I_simulation}

}\subfloat[Magnified View for Case 1]{\includegraphics[width=0.35\linewidth]{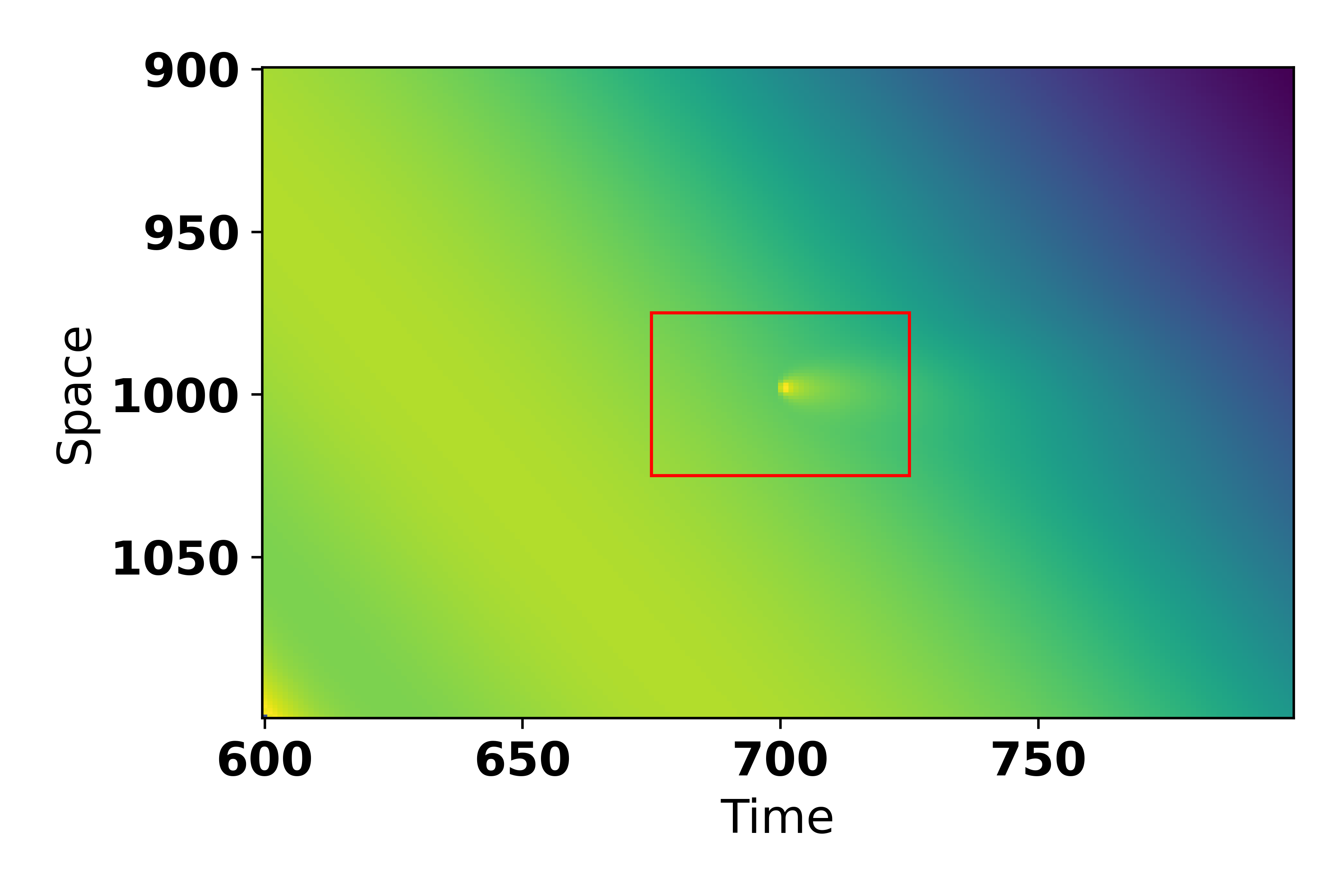}\label{case_I_anomaly}

}

\centering \subfloat[Spatio-temporal Anomaly for Case 1]{\includegraphics[width=0.35\linewidth]{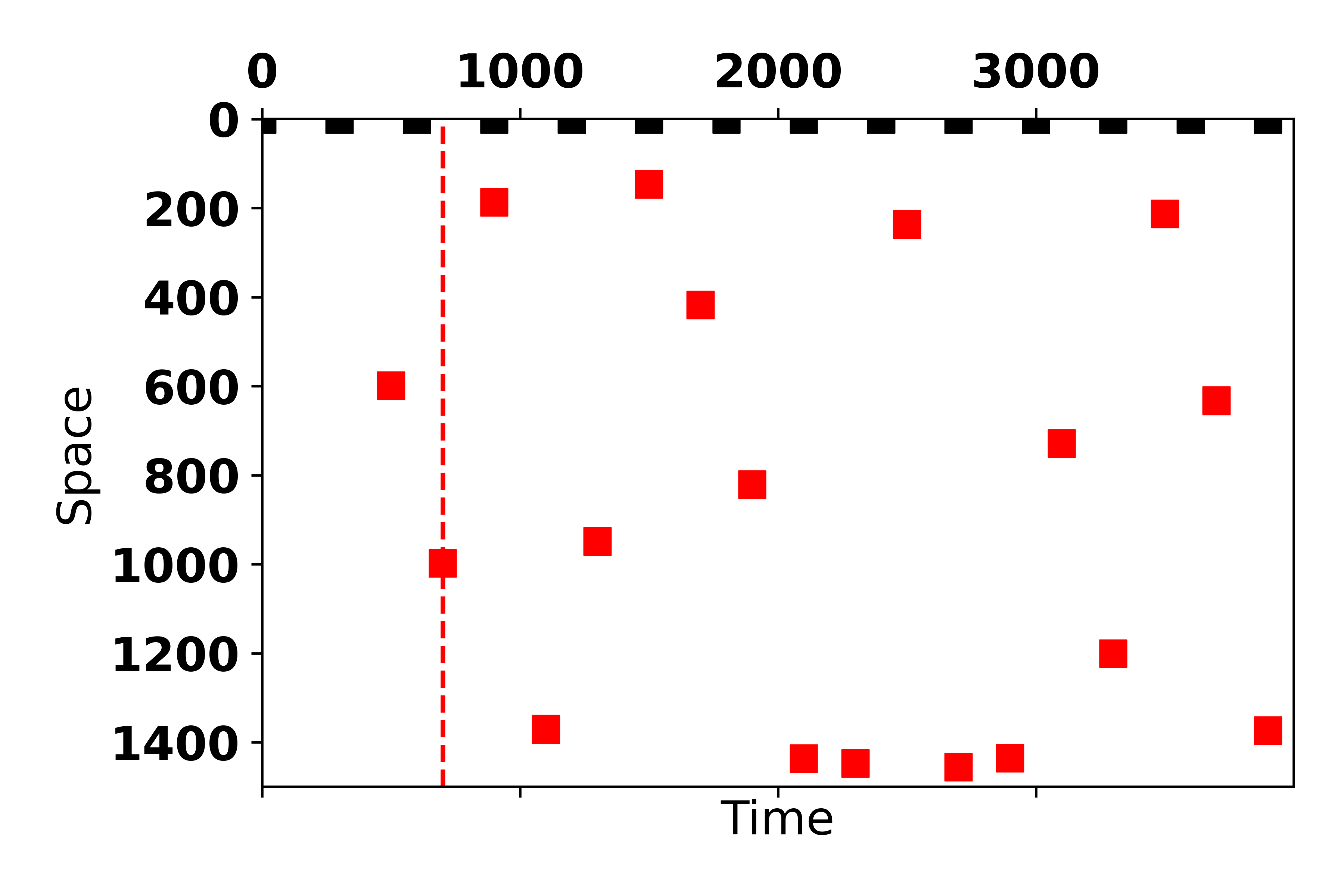} \label{case_I_Stimulation}

}\subfloat[Spatial Map for Case 1]{\includegraphics[width=0.35\linewidth]{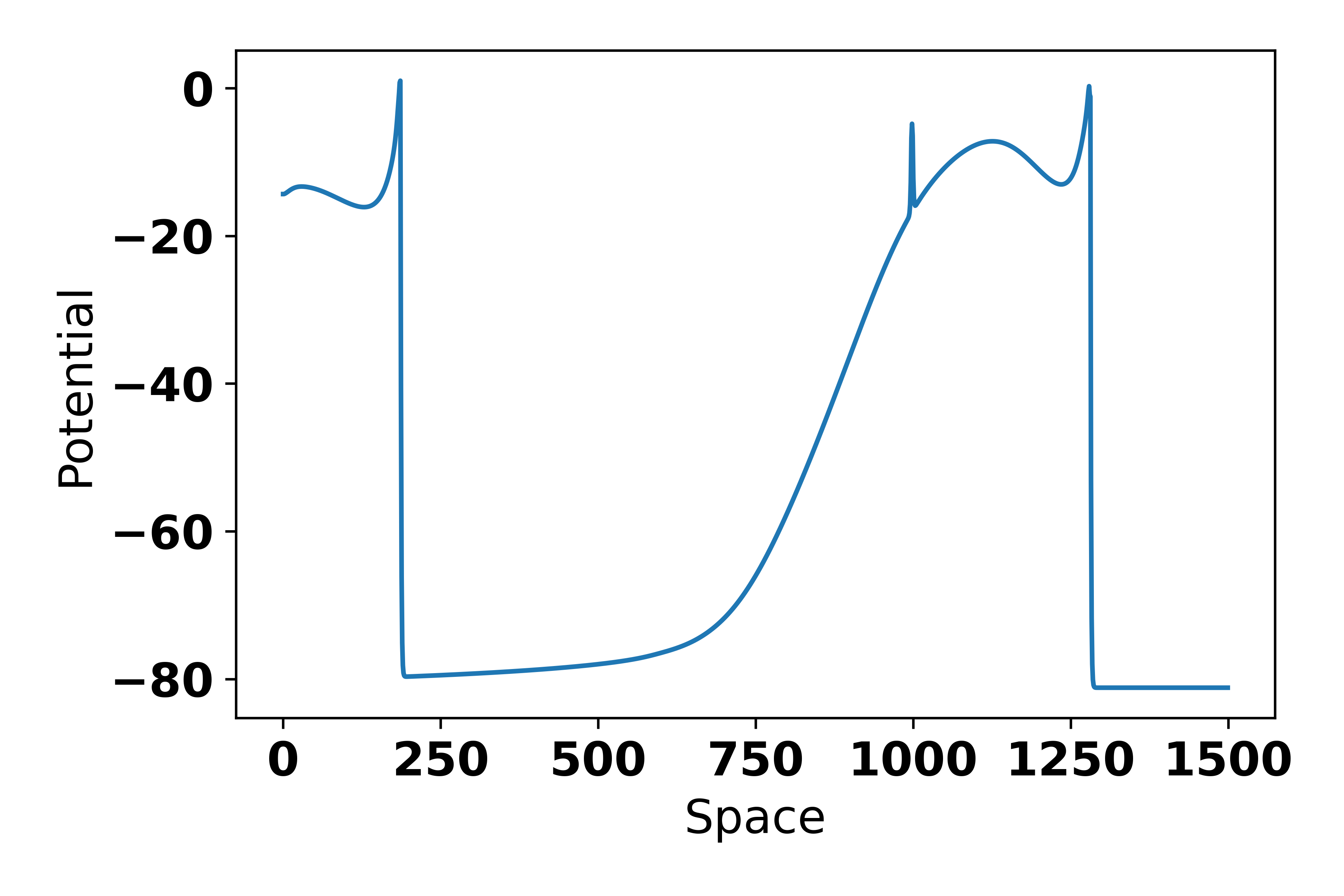}\label{case_I_2d}

}\caption{Here shows the data generated by
the CRN model with a single stimulation at the $1^{\mathrm{st}}$ cell. The X
axis and Y axis in Fig.~\ref{case_I_simulation},Fig.~\ref{case_I_Stimulation} and
Fig.~\ref{case_I_anomaly} represent time and space (cell) index correspondingly. The X axis and Y axis in Fig.~\ref{case_I_2d} represent space(cell) index and the magnitude of cell potential.  Fig.~\ref{case_I_simulation}
describes the raw signal in a 2D plot. It shows the magnitude of the potential
at different spatial-temporal locations. Fig.~\ref{case_I_2d} is an example of
potential along with the cells at $t=700$. Fig.~\ref{case_I_Stimulation}
describes the stimulation where regular stimulation is marked in black, and
abnormal stimulation is marked in red. For Case I, there are repeated normal
stimulation at the 1$^{\mathrm{st}}$ cell and the abnormal stimulation happens
randomly along the cells. Fig.~\ref{case_I_anomaly} visualize one example of
abnormal stimulation at $t=700$ }
\label{Fig: data1} 
\end{figure}

\begin{figure}[h!]
\centering \subfloat[Original Spatio-temporal Map for Case 2]{\includegraphics[width=0.35\linewidth]{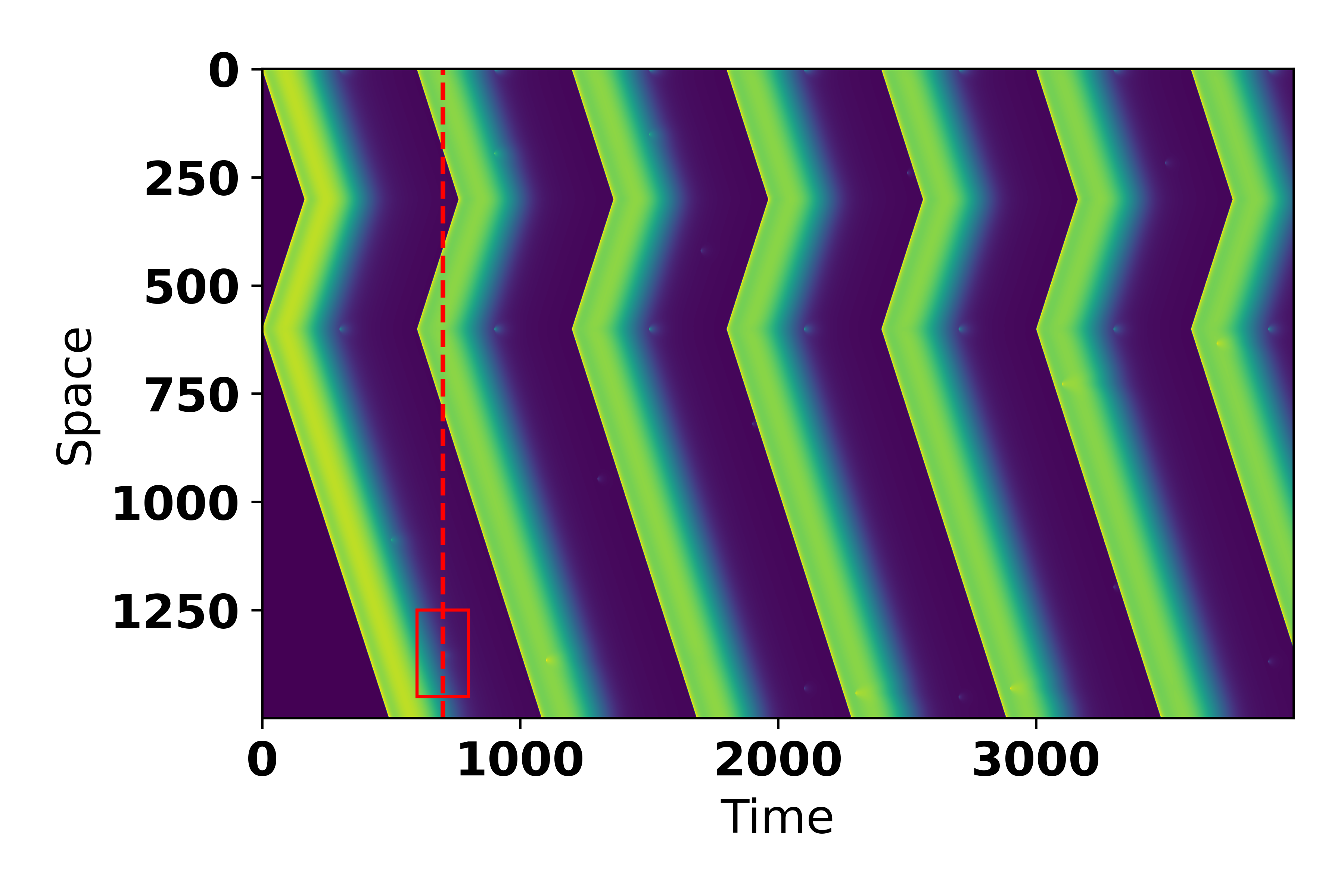} \label{case_II_simulation}

}\subfloat[Magnified View for Case 2]{\includegraphics[width=0.35\linewidth]{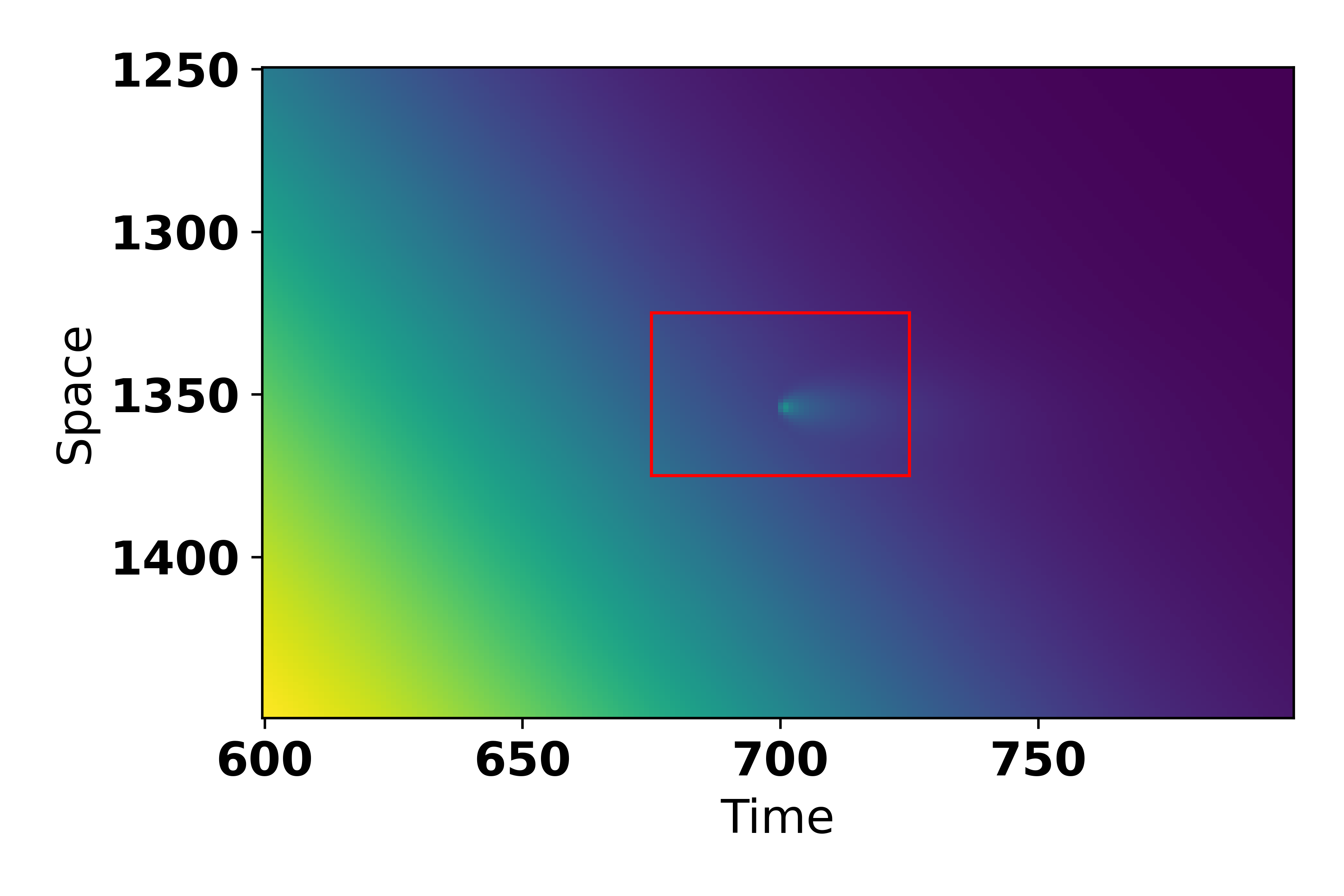}\label{case_II_anomaly}

}

\centering \subfloat[Spatio-temporal Anomaly for Case 2]{\includegraphics[width=0.35\linewidth]{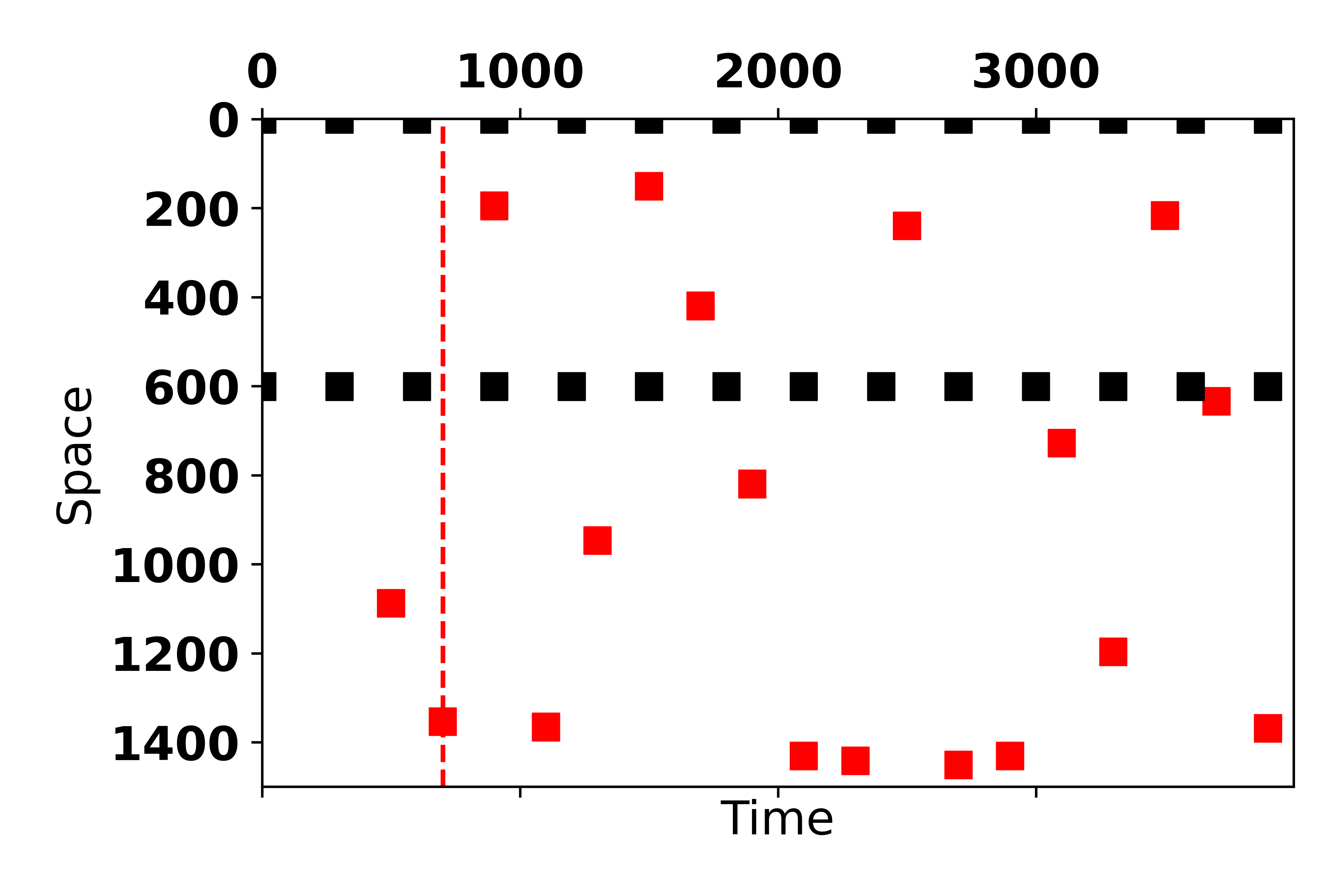} \label{case_II_Stimulation}

}\subfloat[Spatial Map for Case 2]{\includegraphics[width=0.35\linewidth]{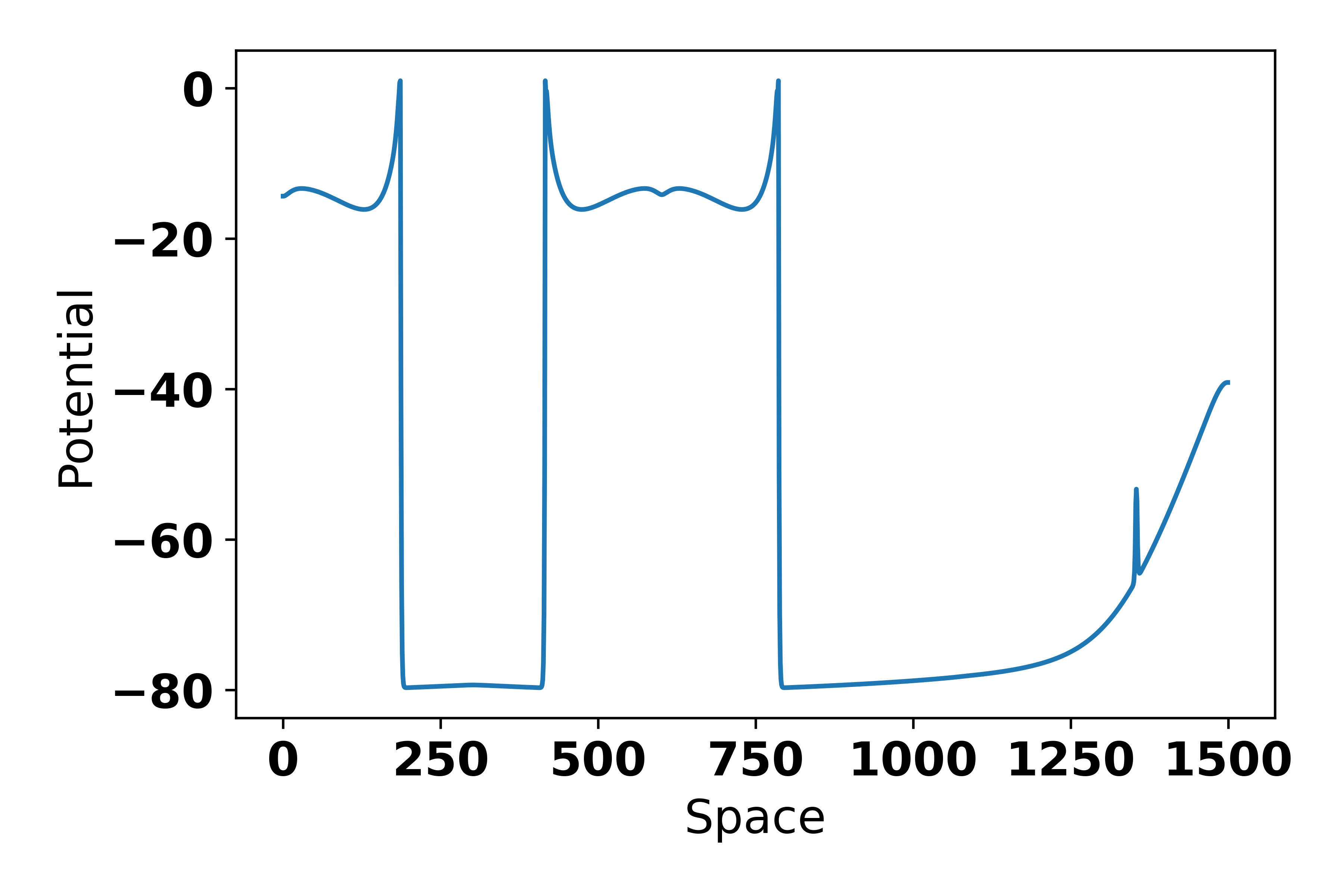}\label{case_II_2d}

}\caption{Here shows the data generated by
the CRN model with two stimulation at the 1$^{\mathrm{st}}$ cell and the 600$^{\mathrm{st}}$ cell. Fig.~\ref{case_II_simulation}
describes the raw signal in a 2D plot. Fig.~\ref{case_II_2d} is an example of potential along with the cells at $t=700$. Fig.~\ref{case_II_Stimulation} describes the stimulation where regular stimulation is marked in black, and abnormal stimulation is marked in red. Fig.~\ref{case_II_anomaly} visualize one example of abnormal stimulation at $t=700$}
\label{Fig: data2} 
\end{figure}

Fig.~\ref{Fig: data1} and Fig.~\ref{Fig: data2}  show two different simulations of the spatio-temporal
propagation on a one-dimensional cell string consisting of 1500
cells. The horizontal axis shows the propagation time in milliseconds, and
the vertical axis shows the cell index. Periodic stimulation is given
to the 1st cell in the first case and to the 1st and 600th cells in
the second case. Fig.~\ref{case_I_simulation} and Fig.~\ref{case_II_simulation} show
the spatio-temporal map of transmembrane potentials. Fig.~\ref{case_I_anomaly} and Fig.~\ref{case_II_anomaly} show the magnified
spatio-temporal 2D map between 600ms and 800ms around the anomaly regions (between the 1250th cell and the 1500th cell). The anomalies at $t=700$ are highlighted by red squares. Further, Fig.~\ref{case_I_Stimulation} and Fig.~\ref{case_II_Stimulation} mark the regular and abnormal stimulation, and Fig.~\ref{case_I_2d} and Fig.~\ref{case_II_2d} show the 1D spatial map around these anomaly regions (i.e., the transmembrane potentials along the cell string).

In both cases, every other stimulation dies out and
does not propagate like others. This is due to the refractoriness
of the cardiac cell, where immediate stimulation after repolarization
within in the cell refractory period cannot be initiated. The refractoriness
of the cardiac cell can change the propagating direction when two
waves merge together, as seen in the right case. It is essential to
model such refractoriness to capture the complex and dynamic activities
of cardiac electrical waves. Besides the regular stimulation, there
are also signals caused by irregular stimulations due to the malfunction
of the cardiac cells.

The goals of this paper are to 1) develop a metamodeling framework
to learn the nonlinear spatio-temporal dynamics from the data/observations
and apply it to predict the spatio-temporal conduction of electrical
waves in future times. Here, the CRN model will be used as a case
study to test and validate the proposed method. However, the proposed
metamodeling framework can be generally applied to other spatio-temporal
systems as well and greatly improve computational efficiency without
losing too much accuracy. 2) given the metamodel, design a real-time
anomaly detection strategy to localize and separate the abnormal stimulation
(i.e., anomaly) automatically from incoming observations.

\section{Methodology\label{Sec: methodology}}

In this subsection, the proposed deep spatio-temporal sparse decomposition
(DSTSD) method is introduced in Section \ref{Sec: DSTSD}. We will
then discuss two deep spatio-temporal architectures that are useful
to model the complicated spatio-temporal structure in Section \ref{sec: muarchitecture},
namely the convolutional WaveNet (Conv-WaveNet), and convolutional
long short-term memory (Conv-LSTM). In addition, to optimize the deep
learning algorithm in the presence of the outlier, we proposed a robust
spatio-temporal learning procedure, which estimates the parameters
of DSTSD in Section \ref{sec: opt}. In Section \ref{sec: longterm},
we discuss how the trained DSTSD can be used to achieve long-term
prediction of the spatio-temporal mean trend. In Section \ref{sec: PhaseIIAnomaly},
we proposed to through solving the inverse problem through a buffer-window
approach to give a more accurate estimation of the anomaly. In Section
\ref{sec: LRT}, the estimated anomaly will be used to conduct a likelihood
ratio test to give an alarm as soon as the anomaly is detected. Finally,
in Section \ref{sec: tunning}, we introduce procedures for the selection
of tuning parameters.

Finally, in this paper, we will use non-bold symbols
to represent scalar, bold symbol $\mathbf{a}$ to represent vectors,
capital bold symbol $\mathbf{A}$ to represent matrices. For a more
detailed notation table, please see the online Appendix. 

\subsection{Deep Spatio-Temporal Sparse Decomposition \label{Sec: DSTSD} }

In this paper, we focus on the modelling of complicated
nonlinear spatio-temporal dynamics in the HD (high dimensional) data
streams. For simplicity, we begin with profile data and suppose a
sequence of profiles $\{\mathbf{y}_{i}\}_{i=1}^{n}$ are available
at $n$ different time instances $\mathbf{\tau}=[\tau_{1},\cdots,\tau_{n}]^{T}$.
For each profile , we assume the observations are taken at the same
group of $p$ spatial locations, denoted as $\mathbf{x}=[x_{1},\cdots,x_{p}]^{T}$.
It is worth noting that the observation locations can vary for different
profiles and it does not introduce any complexity to the implementation
of the proposed method. Then, for the $i^{th}$ profile, we have $\mathbf{y}_{i}=[y(\tau_{i},x_{1}),\cdots,y(\tau_{i},x_{p})]^{T}$.
In addition, the $i^{th}$ profile is assumed to be decomposed as
$\mathbf{y}_{i}=\boldsymbol{\mu}_{i}+\mathbf{e}_{i}$, where $\boldsymbol{\mu}_{i}=[\mu(\tau_{i},x_{1}),\cdots,\mu(\tau_{i},x_{p})]^{T}$
is the mean trend and $\mathbf{e}_{i}=[e(\tau_{i},x_{1}),\cdots,e(\tau_{i},x_{p})]^{T}$
represents the observation noise. Thus, the objective of this research
is to infer the dynamics governing $\mu(t,s)$ for long-term prediction
and anomaly detection. However, due to the complexity of the underlying
mechanism and the scarcity of the available information, this is difficult
to be accomplished using basic modelling approaches. For example,
suppose we are interested in estimating the dynamics of transmembrane
potential, that is, $\mu(t,s)=u(t,s)$. As described in section \ref{Sec: Motivating Study},
$u(t,s)$ is the only variable that can be observed, whose evolution
is collectively regulated by a bunch of unobserved variables, i.e.,
$I_{ion}$ and $\mathbf{v}(t,s)$, through lots of nonlinear equations.
These special features complicate the estimation of the underlying
dynamics of $u(t,s)$. Therefore, we introduce a deep spatio-temporal
sparse decomposition (DSTSD) structure to achieve the research goal.

To learn the complex dynamics, the evolution of $\mu(t,s)$
at location $x_{q}$ is assumed to be governed by the following equation:
\begin{align}
\mu(\tau_{i+1},x_{q}) & =\mu(\tau_{i},x_{q})+f\left(\{\boldsymbol{\mu}_{j}\}_{j\leq i}\right)+\mathbf{c}_{i+1}.\label{eq: eularpde}
\end{align}
This assumption indicates that the difference of the mean trend at
a specific location, i.e., $\mu(\tau_{i+1},x_{q})-\mu(\tau_{i},x_{q})$,
is controlled by two components. The first component $f(\cdot)$ is
a complicated function to be modeled using a neural network, which
takes the historical trajectories at all $p$ locations as inputs.
Introducing the mean at other locations is to model the interaction
among different spatial locations. The intuition for incorporating
historical information of $\mu(t,s)$ in the dynamics is based on
the observation of the CRN model. As mentioned in section \ref{Sec: Motivating Study},
the unobservable take into effect the transmembrane potential
within different time periods. Thus, the information within the former
can be extracted from the historical information of the latter,
i.e., $\{\boldsymbol{\mu}_{j}\}_{j\leq i}$. The second component
$\mathbf{c}_{i}=[c(\tau_{i},x_{1}),\cdots,c(\tau_{i},x_{p})]^{T}$
is the external stimulation exerted on the system. To detect the anomaly,
we assume that the stimulation at time $\tau_{i}$ can be decoupled
as $\mathbf{c}_{i}=\mathbf{r}_{i}+\mathbf{a}_{i}$, where $\mathbf{r}_{i}$
is the regular stimulation as shown in black in Fig. \ref{case_I_Stimulation},
and $\mathbf{a}_{i}$ is the abnormal stimulation as shown in red
in Fig. \ref{case_I_Stimulation}.

In reality, we can not directly use (\ref{eq: eularpde})
to estimate $\mu_{i+1}$ given $\{\mu_{j}\}_{j\leq i}$ for the following
two reasons: 1) the system dynamic $f(\cdot)$ is often unknown. 2)
We only have noisy observations of $\{\boldsymbol{\mu}_{i}\}_{i=1}^{n}$,
i.e., $\{\boldsymbol{y}_{i}\}_{i=1}^{n}$. To estimate $\{\boldsymbol{\mu}_{i}\}_{i=1}^{n}$
given $\{\boldsymbol{y}_{i}\}_{i=1}^{n}$ without knowing the system
dynamics $f(\cdot)$, we propose another function $g(\cdot)$, which
can take the noisy observation $\{\boldsymbol{y}_{i}\}_{j\leq i}$
as input to estimate the system dynamics. To put it simply, $g(\cdot)$
estimate the differences of $\boldsymbol{\mu}_{i+1}-\boldsymbol{\mu}_{i}$.
In addition, to better capture the abnormal stimulation, a spatial
basis $\mathbf{B}_{a}$ is assumed to be existed to decompose $\mathbf{a}_{i}$
as $\mathbf{a}_{i}=\mathbf{B}_{a}\boldsymbol{\theta}_{a,i}$, where
$\boldsymbol{\theta}_{a,i}$ are the temporal coefficients of the
anomaly at time $\tau_{i}$. The spatial basis of the anomaly should represent the spatial structure of the anomaly. Here, we assume that the anomaly is a local clustered region, therefore, a spline basis is used. To put things together, we can have (\ref{eq: ymugae}):

\begin{equation}
\mathbf{y}_{i+1}=\boldsymbol{\mu}_{i}+g\left(\{\mathbf{y}_{j}\}_{j\leq i};\boldsymbol{\theta}\right)+\mathbf{r}_{i+1}+\mathbf{B}_{a}\boldsymbol{\theta}_{a,i+1}+\mathbf{e}_{i+1}.\label{eq: ymugae}
\end{equation}
where $\boldsymbol{\theta}$ is the parameters of function $g(\cdot)$.
Although the proposed structure is motivated by the CRN model, we
would like to emphasize that it is very general and can be applied
for other spatio-temporal dynamics.

To estimate $\boldsymbol{\theta}$, and $\boldsymbol{\theta}_{a,i}$,
we propose a penalized regression model to estimate all the parameters
through the following loss function 
\begin{align}
\begin{array}{c}
l\left(\boldsymbol{\theta},\boldsymbol{\theta}_{a,i+1}\right)=\sum_{i=1}^{n}\parallel\mathbf{y}_{i+1}-g\left(\{\mathbf{y_{j}}\}_{j\leq t};\boldsymbol{\theta}\right)-\boldsymbol{\mu}_{i}\\
-\mathbf{B}_{a}\boldsymbol{\theta}_{a,i+1}-\mathbf{r}_{i+1}\parallel^{2}\\
+\lambda\boldsymbol{\mu}_{i}^{T}\mathbf{R}\boldsymbol{\mu}_{i}+\gamma\|\boldsymbol{\theta}_{a,i+1}\|_{1}\:,
\end{array}\label{eq: decomp}
\end{align}
where $\|\cdot\|_{1}$ is the $L_{1}$ norm operator, and $\lambda$
and $\gamma$ is the tuning parameter to be determined by the user.
$\lambda\boldsymbol{\mu}_{i}^{T}\mathbf{R}\boldsymbol{\mu}_{i}$ encourages
the smoothness of background and $\gamma|\boldsymbol{\theta}_{a,i}|_{1}$
encourage the sparsity of the anomalous regions. The Matrix $\mathbf{R}$
is the regularization matrix that controls the smoothness of the mean
function $\boldsymbol{\mu}_{i}$. For example, one popular choice
for $\mathbf{R}$ is that $\mathbf{R}=\mathbf{D}^{T}\mathbf{D}$,
where $\mathbf{D}$ is the second-order differential operator as $\mathbf{D}=\left[\begin{array}{cccc}
1 & -2 & 1\\
 & \ddots & \ddots\\
 & 1 & -2 & 1
\end{array}\right]$.

Therefore, in the following chapters, we will first discuss two variants
of the spatio-temporal model architectures $g(\cdot)$ for the spatial-temporal
mean trend of the functions and then discuss how to estimate $\boldsymbol{\mu}_{i}$
in the current framework.

\subsection{Spatio-temporal Model Architectures for $\boldsymbol{\mu}_{t}$ \label{sec: muarchitecture}}

In this subsection, we will evaluate two popular deep
learning architectures for the complex spatio-temporal dynamic models
$g\left(\{\mathbf{y}_{j}\}_{j\leq i};\boldsymbol{\theta}\right)$
in (\ref{eq: ymugae}), inspired by the CRN equation in (\ref{eq: CRN_mu}).
In literature, there are many spatio-temporal models that can be used.
Here, we are specifically interested in nonlinear methods with long-term
prediction capacity. We will also evaluate which model is able to
predict the refractoriness effect of cardiac cells. More specifically,
we will focus on two specific models Convolutional WaveNet (Conv-WaveNet)
and Convolutional Long Short-Term Memory Networks Model (Conv-LSTM)
due to their ability to model the long-term dependency. More details
about the specific architecture are discussed in the supplementary material.

\subsubsection{Convolutional WaveNet (Conv-WaveNet) \label{subsubsec: Conv-Wavenet}}

WaveNet was originally introduced to model and generate realistic
audio waveforms by considering the long-term dependency of the time
sequence by the use of deep dilated convolution to increase the receptive
field to model long-term dependency. We propose to extend the WaveNet
architecture with spatial convolution such that the complex spatial
correlation and long-term dependency can be modeled simultaneously\cite{oord2016wavenet}.
Here, we denote the size of the receptive window as $w_{r}$. 
\[
g\left(\{\mathbf{y}_{j}\}_{j\leq t};\boldsymbol{\theta}\right)=g\left(\mathbf{y}_{t-w_{r}:t};\boldsymbol{\theta}\right)
\]
Therefore, $g\left(\mathbf{y}_{t-w_{r}:t};\boldsymbol{\theta}\right)$
is a function of $\mathbf{y}_{t-w_{r}},\mathbf{y}_{t-w_{r}+1},\cdots,\mathbf{y}_{t}$.
Here, the receptive field $w_{r}=2^{d}$, where $d$ is the number
of dilated convolutional layers in the deep neural network. The benefit
of using the WaveNet architecture is that the receptive field increases
exponentially with the depth so that the long-term dependency can
be modeled. More details about the Conv-WaveNet architecture are discussed
in the supplementary material.

\subsubsection{Convolutional Long Short-Term Memory Networks Model (Conv-LSTM) \label{subsubsec: Conv-LSTM}}

LSTM is one type of recurrent neural network that is designed to learn
the long-term dependencies. They are widely used in a large variety
of problems, such as time-series prediction and natural language processing.
However, the LSTM method is not suitable to model the spatio-temporal
propagation since it uses the fully connected transition matrices
on the hidden state, which cannot take advantage of the spatial neighborhood
structure during the temporal transition and could potentially lead
to the overfitting \cite{RN10}. In contrast, Conv-LSTM
is proposed in \cite{RN10} to model this local propagation via the
convolutional operator. We use $\mathbf{z}_{h_{t}}$ to represent
the memory state of Conv-LSTM at time $t$ and Conv-LSTM is a recursive
function to link the data and previous memory state as: 
\begin{align*}
\mathbf{z}_{f_{t}} & =\sigma_{g}(\boldsymbol{\theta}_{W_{f}}*\boldsymbol{\mu}_{t}+\boldsymbol{\theta}_{U_{f}}*\mathbf{z}_{h_{t-1}}+\boldsymbol{\theta}_{V_{f}}\circ\mathbf{z}_{c_{t-1}}+b_{f})\\
\mathbf{z}_{i_{t}} & =\sigma_{g}(\boldsymbol{\theta}_{W_{i}}*\boldsymbol{\mu}_{t}+\boldsymbol{\theta}_{U_{i}}*\mathbf{z}_{h_{t-1}}+\boldsymbol{\theta}_{V_{i}}\circ\mathbf{z}_{c_{t-1}}+b_{i})\\
\mathbf{z}_{c_{t}} & =\mathbf{z}_{f_{t}}\circ\mathbf{z}_{c_{t-1}}+\mathbf{z}_{i_{t}}\circ\sigma_{c}(\boldsymbol{\theta}_{W_{c}}*\boldsymbol{\mu}_{t}+\boldsymbol{\theta}_{U_{c}}*\mathbf{z}_{h_{t-1}}+b_{c})\\
\mathbf{z}_{o_{t}} & =\sigma_{g}(\boldsymbol{\theta}_{W_{o}}*\boldsymbol{\mu}_{t}+\boldsymbol{\theta}_{U_{o}}*\mathbf{z}_{h_{t-1}}+\boldsymbol{\theta}_{V_{o}}\circ\mathbf{z}_{c_{t-1}}+b_{o})\\
\mathbf{z}_{h_{t}} & =\mathbf{z}_{o_{t}}\circ\sigma_{h}(\mathbf{z}_{c_{t}})\\
\boldsymbol{\mu}_{t+1} & =\boldsymbol{\mu}_{t}+\mathbf{z}_{h_{t}}+\mathbf{c}_{t+1}
\end{align*}
Again, motivated by the Euler's equation, the LSTM model is used to
model the difference between $\mathbf{\mu_{t}}$ and $\mathbf{\mu_{t+1}}$.
Here, we use $\mathbf{z}_{f_{t}},\mathbf{z}_{i_{t}},\mathbf{z}_{c_{t}},\mathbf{z}_{o_{t}},\mathbf{z}_{h_{t}}$
to denote the latent state variables, namely the forget gate, input gate, cell state, output gate, and hidden state inside the LSTM model, and we use $\splitatcommas{\boldsymbol{\theta}_{W_{f}},\boldsymbol{\theta}_{U_{f}},\boldsymbol{\theta}_{V_{f}},\boldsymbol{\theta}_{W_{i}},\boldsymbol{\theta}_{U_{i}},\boldsymbol{\theta}_{V_{i}},\boldsymbol{\theta}_{W_{c}},\boldsymbol{\theta}_{U_{c}},\boldsymbol{\theta}_{V_{c}},\boldsymbol{\theta}_{W_{o}},\boldsymbol{\theta}_{U_{o}},\boldsymbol{\theta}_{V_{o}}}$
to denote the parameters for LSTM model \cite{RN10}, which are the parameters for the forget gate, input gate, cell state, and output gate, respectively. Notation $\circ$
represents the Hadamard product and $*$ represents the convolution
operator. 

\subsection{Phase-I Analysis \label{sec: opt}}

In the Phase-I analysis, we will discuss the algorithm to optimize
$\boldsymbol{\theta}$ and $\boldsymbol{\theta}_{a,t}$ in the off-line
setting for Phase-I analysis. We assume that a set of spatio-temporal
data $y_{i,t}$ with length $N_{t}$ will be collected with $i=1,\cdots,N$.
To simplify the cases, we assume that the outliers in Phase-I analysis,
if exist, are often random, which corresponds to $\mathbf{B}_{a}=\mathbf{I}$.

We first prove that solving $\boldsymbol{\theta}$ and $\boldsymbol{\theta}_{a}$
in (\ref{eq: decomp}) is equivalent to optimize the $\boldsymbol{\theta}$
with the Huber loss function in the following proposition and then
the soft thresholding on the residual. 
\begin{prop}
When $\mathbf{B}_{a}=\mathbf{I}$, in (\ref{eq: decomp}), $\boldsymbol{\theta}$
can be solved by 
\begin{align}
\boldsymbol{\theta} & =\arg\min_{\boldsymbol{\theta}}l_{r}\left(\boldsymbol{\theta}\right)\label{eq: robusttheta}\\
\boldsymbol{\theta}_{a,i,t+1} & =S_{\gamma/2}\left(\mathbf{y}_{i,t+1}-g\left(\{\mathbf{y}_{i,t'}\}_{t'\leq t};\boldsymbol{\theta}\right)-\boldsymbol{\mu}_{i,t}-\mathbf{r}_{t+1}\right),\label{eq: robustthetaa}
\end{align}
where $l_{r}\left(\boldsymbol{\theta}\right)$ is defined as: 
\begin{align}
\begin{array}{c}
l_{r}\left(\boldsymbol{\theta}\right)=\sum_{i=1}^{N}\sum_{t=1}^{N_{t}}(\rho(\mathbf{y}_{t+1}-g\left(\{\mathbf{y}_{i}\}_{i\leq t};\boldsymbol{\theta}\right)-\boldsymbol{\mu}_{t}-\mathbf{r}_{t+1})+\\
\lambda\boldsymbol{\mu}_{t}^{T}\mathbf{R}\boldsymbol{\mu}_{t}).
\end{array}\label{eq:  Robustl}
\end{align}

Here, $\rho(x)$ is the Huber loss function, defined by $\rho(x)=\begin{cases}
x^{2} & |x|\leq\frac{\gamma}{2}\\
\gamma|x|-\frac{\gamma^{2}}{4} & |x|>\frac{\gamma}{2}
\end{cases}$. $S_{\gamma}(x)=\mathrm{sgn}(x)(\left|x\right|-\gamma)_{+}$ is the
soft thresholding operator, in which $\mathrm{sgn}(x)$ is the sign
function and $x_{+}=\max(x,0)$. 
\end{prop}

The proof is given in the Supplementary Material.

Finally, given the loss function in \eqref{eq:  Robustl}, the parameter
$\boldsymbol{\theta}$ can be solved by the combination of the back-propagation
and the stochastic gradient descent to update the model parameter
$\boldsymbol{\theta}$ based on a mini-batch of samples in the $k^{th}$
iteration. More specifically, in the Conv-WaveNet model, since $g\left(\{\mathbf{y}_{i}\}_{i<t};\boldsymbol{\theta}\right)=g(\mathbf{y}_{t-w_{r}:t};\boldsymbol{\theta})$
, $\frac{g\left(\mathbf{y}_{t};\boldsymbol{\theta}\right)}{\partial\boldsymbol{\theta}}$
can be directly computed via the back-propagation. However, for the
Conv-LSTM model, the gradient will flow back into the starting time,
which increases the computational complexity dramatically for large
$t$. Normally, truncated back-propagation can be applied to cut the
gradient flow in the latest few measurements to decrease the computational
complexity.

\subsection{Real-time Long-term Prediction \label{sec: longterm}}

The previous subsection focuses on training the spatio-temporal models
in the off-line setting. However, since the temporal dimension is
changing over time, it is not trivial to apply the model in the online-setting
for real-time long-term prediction. In the example
of the cardiac electric conduction, it is important
to predict the future events in a couple of cardiac cycles (i.e, heartbeats)
for over 500ms to 1000ms ($\Delta t$ = 0.1ms). In the anomaly detection
application, the long-term prediction provides references to identify
abnormal stimulation. To achieve this, we will discuss how to apply
the trained model in the online setting in real-time.

In this subsection, we will discuss how to enable the long-term prediction
for both Conv-WaveNet models and Conv-LSTM models. However, since
the temporal dependency of these two models is different, we will
discuss them separately as follows:

\textbf{\textit{Conv-WaveNet:}} We will discuss how to enable the
long-term prediction for the Conv-WaveNet model. Here, we denote $\boldsymbol{\mu}_{{t_{0}}}(t_{0}+\Delta t)$
as the $\Delta t$-ahead prediction of $\boldsymbol{\mu}(t_{0}+\Delta t)$
at time $t_{0}$. For the long-term prediction, the following method
can be used. We know that $\mathbf{\hat{\boldsymbol{\mu}}}\left(t_{0}+i+1\right)=\mathbf{c}_{{t_{0}+i+1}}+\hat{\boldsymbol{\mu}}\left(t_{0}+i\right)+g\left(\mathbf{y}_{{t_{0}+i-w_{r}:t_{0}+i}};\boldsymbol{\theta}\right)$.
Therefore, we can derive the following formula for the long-term prediction.
\[
\mathbf{\hat{\boldsymbol{\mu}}}_{{t_{0}}}\left(t_{0}+\Delta t\right)=\mathbf{y}_{t_{0}}+\sum_{i=1}^{\Delta t-1}g\left(\mathbf{\hat{\boldsymbol{\mu}}}_{{t_{0}}+i-w_{r}:{t_{0}}+i};\boldsymbol{\theta}\right)+\mathbf{c}_{{t_{0}}+\Delta t}
\]
Here, typically, in the real-time prediction, we will set the future
anomaly $\mathbf{a}_{t}=0$. However, in some rare cases, the future
anomaly source is already known $\mathbf{a}_{t}$, this method can
also predict how the system reacts to the anomaly accurately.

\textbf{\textit{Conv-LSTM:}} Similarly, we would like to discuss how
to enable long-term prediction for the Conv-LSTM model. Unlike the
Conv-WaveNet model, the predicted value $\hat{\mu}_{{t_{0}}}(t_{0}+1)$
requires all values $\mathbf{y}_{t},t=1,\cdots t_{0}$. We divide
the long-term prediction into two phases: the warm-up phase and the
prediction phase. In the warm-up phase, we will start with $y_{0}$
or some value from $\mathbf{y}_{{t^{'}}},t^{'}<t_{0}$ to learn a
more accurate memory state representation $h_{t}$ from the original
data as well as estimating the mean trend $\boldsymbol{\mu}_{t}$
in the past. In the warm-up phase, the $\mathbf{y}_{t}$ is known
for $t=1,\cdots t_{0}$, therefore, $\mathbf{y}_{t}$ can be used
as input for the Conv-LSTM model. Furthermore, $\mathbf{\hat{a}}_{t+1}=\hat{\boldsymbol{\theta}}_{a,t+1}$
in the phase-I analysis, can be estimated by (\ref{eq: robustthetaa}).
In the prediction phase, $\mathbf{y}_{t}$ is not known for $t>t_{0}$.
In this case, we propose to use the future prediction $\mathbf{\hat{\mu}}_{t}$
for $t>t_{0}$. In the long-term prediction phase, if we know the
future stimulation $\mathbf{a}_{t}$, this can be combined in future
prediction. If we do not know where and when the future stimulation
is, we typically set $\mathbf{a}_{t}=0$.

\subsection{Online Anomaly Estimation \label{sec: PhaseIIAnomaly}}

In this subsection, we will discuss how to apply the proposed algorithm
for online anomaly detection. More specifically, we assume that the
anomaly is sparse and only happens at a certain time interval $t\in\left[T_{0},T_{0}+w\right]$
(i.e., epidemic change \cite{ravckauskas2004holder}). This type of
change is very common in the cardiac electrical conduction. Mathematically
speaking, we define the normal and abnormal transition in (\ref{eq: epidemicchange}).

\begin{align}
\boldsymbol{\mu}_{t+1}\left(s\right)=\boldsymbol{\mu}_{t}\left(s\right)+f\left(\boldsymbol{\mu}_{t}\left(s\right)\right)+\mathbf{r}_{t+1}, & t<T_{0}\text{ or }t>T_{0}+T_{w}\nonumber \\
\boldsymbol{\mu}_{t+1}\left(s\right)=\boldsymbol{\mu}_{t}\left(s\right)+f\left(\boldsymbol{\mu}_{t}\left(s\right)\right)+\mathbf{r}_{t+1}+\mathbf{a}_{t+1}, & t\in[T_{0},T_{0}+T_{w}],\label{eq: epidemicchange}
\end{align}

Detecting the epidemic change is very challenging. The reason is that
if we design a control chart methodology only based on the Q-control
chart, designed based on the model residual such as $Q(t)=\|\mathbf{y}_{t}-\boldsymbol{\mu}_{t-1}-g\left(\{\mathbf{y}_{t'}\}_{t'<t-1};\boldsymbol{\theta}\right)\|^{2}$,
$Q(t)$ will be small for $t<T_{0}$ or $t>T_{0}+T_{w}$ and only
be large during the epidemic change window $[T_{0},T_{0}+T_{w}]$.
Therefore, if the algorithm fails to detect the anomaly at time $t\in[T_{0},T_{0}+T_{w}]$,
it may never detect the anomaly again in the future time $t'>T_{0}+T_{w}$
since the anomaly will be combined into the future spatio-temporal
mean trend in the next time as $\boldsymbol{\mu}_{t}=\boldsymbol{\mu}_{t-1}+g\left(\{\mathbf{y}_{t'}\}_{t'<t-1};\boldsymbol{\theta}\right)+\mathbf{a}_{t}$
and results in small residual $Q(t)$ for future $t'>T_{0}+T_{w}$.

Another aspect is that when the anomaly happens at time $T_{0}$,
it will start with a small magnitude at $T_{0}$ and then propagate
to a large area in the future time $t>T_{0}$. Therefore, it is often
much effective to detect such change from a retrospective perspective
to analyze the change point and location that may happen in the past.
However, a full perspective requires scanning all possible locations
of changes back in time, which is computationally inefficient.

To address this, we propose to use a buffer window to provide a better
estimation of the anomaly event. Suppose we would like to detect change
at time $T$, we propose to use $T+w$ to $T$ as a buffer period
to estimate the source of the anomaly. This may naturally introduce
a detection delay due to the buffer window $w$ but will create a
better estimation of the anomaly. For a special case, $w=0$, only
data $y_{T}$ will be used to detect the change at time $T$. For
more discussion about choosing the best buffer period, please refer
to Section \ref{sec: tunning}.

This procedure relies on the long-term prediction capacity of the
proposed algorithm. For example, we assume that under this buffer
period, the true data $\mathbf{y_{t}}$ is not measured. Therefore,
to estimate the change, we have to rely on the predicted $\mathbf{\mu_{t}}$
when $t>T$. Therefore, we have:

\begin{align}
\boldsymbol{\mu}_{t+k} & =\boldsymbol{\mu}_{t+k-1}+g\left(\{\boldsymbol{\mu}_{t'}\}_{t\leq t'<t+k};\boldsymbol{\theta}\right)+\mathbf{a}_{t+k}+\mathbf{r}_{t+k}\label{eq: mu_recursiwve}\\
 & s.t.\quad k=1,\cdots,w.\nonumber 
\end{align}
Furthermore, in Phase-II monitoring, we assume that the spatio-temporal
model has been trained before, and $\boldsymbol{\theta}$ has to be
estimated. We will rely on the following optimization algorithms to
estimate the $\boldsymbol{\theta}_{a,t}$. Here, one specific challenge
is the recursive formula of $g$. Since $\boldsymbol{\mu}_{t}$ relies
on $\boldsymbol{\mu}_{t-1}$, which in turns relies on $\mathbf{a}_{t-1}$.
Therefore, the following loss function aims to optimize
or estimate the anomaly from time $T$ to $T+w$, namely $\mathbf{a}_{T}$
to $\mathbf{a}_{T+w}$.  The challenge is that the problem is highly
coupled, given the recursion of $\boldsymbol{\mu}_{t+k}$, as shown
in (\ref{eq: RegWindowLoss}). 
\begin{align}
\min_{\{\boldsymbol{\theta}_{a,t}\}_{t\in[T,T+w]}}\sum_{t=T}^{T+w} & \|\mathbf{y}_{t}-\boldsymbol{\mu}_{t}\|^{2}+\gamma\sum_{t=T}^{T+w}\|\boldsymbol{\theta}_{a,t}\|_{1}\label{eq: RegWindowLoss}\\
s.t.\quad\boldsymbol{\mu}_{T+w} & =\boldsymbol{\mu}_{T+w-1}+g\left(\{\boldsymbol{\mu}_{t'}\}_{t'<t+k};\boldsymbol{\theta}\right)+\\
 & \mathbf{B}_{a}\boldsymbol{\theta}_{a,T+w}+\mathbf{r}_{t+w}\nonumber \\
 & \vdots\nonumber \\
\quad\boldsymbol{\mu}_{T} & =\boldsymbol{\mu}_{T-1}+g\left(\{\boldsymbol{\mu}_{t'}\}_{t'<T};\boldsymbol{\theta}\right)+\mathbf{B}_{a}\boldsymbol{\theta}_{a,T}+\mathbf{r}_{T}.\nonumber 
\end{align}

To minimize the regularized loss function in (\ref{eq: RegWindowLoss}),
we propose to first plug in all the $\mathbf{\mu_{t}}$ into the definition
of the $\boldsymbol{\mu}_{t+1}$ for $t=T,\cdots T+w$ as a function
$\boldsymbol{\mu}_{t}=\boldsymbol{\mu}_{t-1}+g\left(\{\boldsymbol{\mu}_{t'}\}_{t'<t};\boldsymbol{\theta}\right)+\mathbf{B}_{a}\boldsymbol{\theta}_{a,t}+\mathbf{r}_{t}=\cdots=\boldsymbol{\mu}_{t}(\{\boldsymbol{\theta}_{a,t'}\}_{t'<t})$.
Finally, we define the windowed loss function as $l_{T\rightarrow T+w}(\{\boldsymbol{\theta}_{a,t}\})$
in (\ref{eq: mu_window}).

\begin{equation}
l_{T\rightarrow T+w}(\{\boldsymbol{\theta}_{a,t}\})=\sum_{t=T}^{T+w}\|\mathbf{y}_{t}-\boldsymbol{\mu}_{t}(\{\boldsymbol{\theta}_{a,t'}\}_{t'<t})\|^{2}\label{eq: mu_window}
\end{equation}
Finally, by plugging in the windowed loss function in (\ref{eq: RegWindowLoss}),
we have (\ref{eq: proxloss}).

\begin{equation}
\min_{\boldsymbol{\theta}_{a,t},t\in[T,T+w]}l_{T\rightarrow T+w}(\{\boldsymbol{\theta}_{a,t}\})+\gamma\sum_{t=T}^{T+w}\|\boldsymbol{\theta}_{a,t}\|_{1}.\label{eq: proxloss}
\end{equation}
Finally, the loss function in (\ref{eq: proxloss}) can be decomposed
into two terms, where $l_{T\rightarrow T+w}(\{\boldsymbol{\theta}_{a,t}\})$
is differentiable and $\sum_{t=T}^{T+w}\|\boldsymbol{\theta}_{a,t}\|_{1}$
is non-differentiable but has a rather simple proximal operator. Therefore,
the proximal gradient algorithm can be used to optimize $\boldsymbol{\theta}_{a,t}$. 
\begin{prop}
In $k^{th}$ iteration, $\boldsymbol{\theta}_{a,t}^{(k+1)}$ in (\ref{eq: proxloss})
can be optimized by 
\begin{equation}
\boldsymbol{\theta}_{a,t}^{(k+1)}=S_{\gamma/2}(\boldsymbol{\theta}_{a,t}^{(k)}-c\frac{\partial}{\partial\boldsymbol{\theta}_{a,t}}l_{T\rightarrow T+w}(\{\boldsymbol{\theta}_{a,t}\}),\label{eq: thetaat_softthreshold}
\end{equation}
where $c$ is the step size of the proximal gradient algorithm and
$S_{\gamma/2}(\cdot)$ is the soft-thresholding operator. 
\end{prop}

The proof is given in Supplementary Material.

It is worth noting that for convex and Lipschitz continuous function
$l_{T\rightarrow T+w}(\cdot)$, $\boldsymbol{\theta}_{a,t}$ will
converge to the global optimum. However, since $l_{T\rightarrow T+w}(\cdot)$
is highly non-convex from the deep learning architectures, it is often
impossible to guarantee the convergence. However, in reality, we find
out with only a few iterations, the algorithm can already obtain a great
estimation of the anomaly $\boldsymbol{\theta}_{a,t}$. Finally, this
estimated $\boldsymbol{\theta}_{a,t}$ will be used to construct the
monitoring statistics, which will be discussed in Section \ref{sec: LRT}.

\subsection{Anomaly Detection Through the Likelihood Ratio Test \label{sec: LRT}}

After $\boldsymbol{\theta}_{a,t}$ has been solved, we will construct
a likelihood ratio test to detect the change over time. We know from
(\ref{eq: ymugae}) that $\mathbf{r}_{t}=\mathbf{y}_{t}-\boldsymbol{\mu}_{t-1}-g\left(\{\mathbf{y}_{i}\}_{i<t};\boldsymbol{\theta}\right)=\mathbf{B}_{a}\boldsymbol{\theta}_{a,t}+\mathbf{e}_{t}.$
If $\boldsymbol{\theta}_{a,t}=0$, there will be no anomaly and $\mathbf{r}_{t}\sim N(0,\sigma^{2}I)$.
If there is an anomaly, $\mathbf{r}_{t}\sim N(\mathbf{B}_{a}\hat{\boldsymbol{\theta}}_{a,t},\sigma^{2}I)$
. Therefore, we can propose a likelihood ratio procedure to test the
mean of $\mathbf{r}_{t}$, denoted as $\mu_{\mathbf{r}_{t}}$ as follows:
\[
H_{0}:\boldsymbol{\mu}_{r_{t}}=0,\quad H_{1}:\boldsymbol{\mu}_{r_{t}}=\mathbf{B}_{a}\boldsymbol{\hat{\theta}}_{a,t}.
\]
Moreover, in this paper, we propose to use a likelihood ratio test
procedure to test whether there is a change in the estimated anomaly
solved by the inverse problem. Finally, according to \cite{wang2009high,zhang2018weakly},
we can derive the following likelihood ratio-test statistics 
\begin{equation}
T_{t}=2\hat{\boldsymbol{\theta}}_{a,t}^{T}\mathbf{B}_{a}^{T}\left(\mathbf{y}_{t+1}-\boldsymbol{\mu}_{t}-g\left(\{\mathbf{y}_{i}\}_{i\leq t};\boldsymbol{\theta}\right)\right)-\hat{\boldsymbol{\theta}}_{a,t}^{T}\mathbf{B}_{a}^{T}\mathbf{B}_{a}\hat{\boldsymbol{\theta}}_{a,t}.\label{eq: Tt}
\end{equation}
Correspondingly, we chose a control limit $L>0$ for (\ref{eq: Tt})
and define if $T_{t}>L$, the monitoring scheme triggers an OC alarm
at time $t$.

\subsection{Tuning Parameter Selection \label{sec: tunning}}

In this subsection, we will discuss the procedure of selecting the
best tuning parameters, including the buffer window size $w$, anomaly
basis $\mathbf{B_{a}}$, control limit $L$, smoothing parameter $\lambda$,
and sparsity parameter $\gamma$.

First, we would like to discuss the procedure of choosing the buffer
window size $w$. In reality, the best $w$ depends on the signal-noise
ratio, defined by the magnitude of the change divided by the noise
magnitude. For a larger signal-noise ratio, it is often easier to
detect and a smaller $w$ is recommended (i.e., $w=0$ ). However,
for a smaller signal-noise ratio, it is often recommended to use a
larger $w$. In reality, it is often hard to know the change magnitude
beforehand. Therefore, we suggest to choose to construct the control
chart with a buffer window from $w={0,1,\cdots,W}$, and select the
one with the smallest detection delay. For example, at time $t$,
we can decide whether time $t-W$ to $t$ has an anomaly due to the use
of different buffer windows. The algorithm will stop until it triggered
the first anomaly.

Second, selecting the anomaly basis is also essential. Selecting a
basis for anomalous regions depends on the type of anomalies we aim
to detect. For example, if anomalies are randomly scattered over the
mean, it is recommended to use an identity basis, i.e., $\mathbf{B}_{a}=I$.
If anomalies form clustered regions, a spline basis or kernel basis
can be used. More details about the spatial basis selection of the
functional mean and anomalies are given in \cite{yan2017anomaly}.

Third, we like to discuss the procedure of choosing the control limit
$L$. Specifically, given a pre-specified IC average run length ($ARL_{0}$),
we propose to select the control limit $L$ by simulation. Given the
complicated spatial-temporal distribution of the data, it is often
hard to get the exact distribution of $T_{t}$. In particular, we
first choose an initial value for $L$, and then compute the $ARL_{0}$
of the monitoring statistic in (\ref{eq: Tt}) based on a large number
of simulation replications, where the IC samples are generated from
the IC distribution of the process. If the computed $ARL_{0}$ is
smaller than the nominal one, we increase the value of $L$. Otherwise,
we decrease it. We repeat this process until the $ARL_{0}$ is achieved
with the desired precision. In particular, in the searching procedure,
we may use some numerical searching algorithms, such as the bisection
search algorithm \cite{qiu2008distribution}.

Forth, we like to discuss the procedure of choosing the smoothing
parameter $\lambda$. Here, $\lambda$ is selected by the cross-validation
procedure, where the validation points are randomly selected points
across the entire sequence $\mathbf{y_{t}}$. The data at the validation
points will be set to $0$. The algorithm will try to recover the
spatio-temporal mean trend on the validation set and compared it with
the original $\mathbf{y_{t}}$.

Finally, we like to discuss the procedure of choosing the sparsity
parameter $\gamma$. In this procedure, $\gamma$ will be selected
based on the fixed false discovery rate as $5\%$ in the Phase-I analysis
with the same basis $\mathbf{B}_{a}$ used in Phase-II analysis. We
will select $\gamma$ such as $5\%$ of the $\boldsymbol{\theta}_{a,t}$
will be detected as an anomaly. The procedure is described in detail
in \cite{yan2019akm2d}.

\section{Simulation Study \label{sec: Simulation Study}}

In this section, we will first discuss the data generation and Experimentation
Details in Section \ref{subsec: generation} and Section \ref{subsec: expsetup}.
The proposed method will be evaluated in terms of metamodeling and
mean-trend prediction in Section \ref{subsubsec: mean-trend} and
anomaly detection in Section \ref{subsubsec: anomaly_detection}.

\subsection{Data Generation \label{subsec: generation}}

We will use the CRN model described in (\ref{eq: CRN_mu})
to perform simulations on a one-dimensional cell array (i.e., 1-D
cable) with 1500 cells and assume the mono-domain tissue model. Samples
(i.e., cell transmembrane potentials) generated by simulations are
used to train the metamodel for the spatio-temporal mean trend in
the proposed DSTSD model. In this study, we are going to consider
two simulation protocols as follows. 
\begin{itemize}
\item \textit{Case 1}: One stimulation at a variable cycle length of 200ms
(5Hz) to 1000ms (1Hz) in a 100ms increment is given to the left end
of the cell array, which triggers electrical waves to propagate to the
other end of the cable. In addition, more experiments were done by
moving the stimulation to the right of the cable in a step of 100
cells. For example, stimulation is given at the left end of the cell
array every 300ms. (See Fig. \ref{case_I_simulation}). 
\item Case 2: Two periodic stimulations at a variable cycle length of 200ms
(5Hz) to 1000ms (1Hz) in a 100ms increment are given at different
locations. The two stimulations are at a variable distance of 300
cells and 600 cells. For example, the stimulations are given at the
1$^{\mathrm{st}}$ cell and the 600th$^{\mathrm{th}}$ cells (see
Fig. \ref{case_II_simulation} and Fig. \ref{case_II_Stimulation}). 
\end{itemize}
In the Phase-II analysis, we still use the two simulation protocols,
Case 1 and Case 2, as described before. We further generate anomalies
on top of the regular stimulation $\mathbf{r_{t}}$, which represents
cell malfunctioning. The abnormal stimulation is randomly picked along
the cell array with an intensity of $\delta$. We design the anomaly
as a sequence of abnormal points, which will cause a continuous stimulation
on the cell. More specifically, we illustrate two cases here with
three consecutive abnormal points, causing stimulation at with the
amplitude ranging from 4.5 to 11 that will last for 2 ms. In another
word, we choose the anomaly $\mathbf{a}_{s,t}=\delta R_{0}1(s\in S_{A})1(t\in S_{T})$
in (\ref{eq: CRN_mu}), where $R_{0}$ is the magnitude of regular
stimulation. $S_{A}$ is the set of anomalous pixels, and $S_{T}$
is the set of time points with anomalies generated. $\delta$ characterizes
the relative intensity difference between anomalies and the spatio-temporal
mean trend. $1(\cdot)$ is the indicator function. In this study,
we choose $S_{A}=\left\{ s_{0},s_{0}+1,s_{0}+2\right\} $ and $S_{T}=\left\{ t_{0},t_{0}+1\right\} $.
Here, the $s_{0}$ and $t_{0}$ are chosen randomly. $t_{0}$ is the
start time of the change. A sample of the simulated spatio-temporal
mean trend and multiple randomly chosen anomalies are shown in Fig.
\ref{case_I_simulation}. In the Phase-I analysis for the training
of the spatio-temporal metamodel, we have generated 15 samples according
to Case 1 and 10 samples according to Case 2. Each sample contains
1000 measurements in 1-ms time step.


\subsection{Experimentation Details \label{subsec: expsetup}}

This subsection will give more details on the experimental details
in both Phase-I and Phase-II analysis.

In the phase I analysis, we first need to fit the cardiac dynamics
with the proposed Conv-LSTM and Conv-WaveNet, which is important to
achieve good performance on long-term prediction and anomaly detection.
Please see the supplementary material for all the details of the training
of Conv-LSTM and Conv-WaveNet.

In the phase-II analysis, a multi-step loss is considered for solving
the inverse problem. The window size is set to 3 regarding the trade-off
between detection delay and detection accuracy. The learning rate
$c=0.01$, and each solving process includes 5 epochs.

\subsection{Result Comparison \label{subsec: result comparison}}

In this subsection, we aim to evaluate the performance of the proposed
method in two different parts. First, we will evaluate the performance
of the proposed algorithm in terms of predicting the spatio-temporal
mean trend in Section \ref{subsubsec: mean-trend}. This also evaluates
the performance of the proposed DSTSD in terms of metamodeling. 2)
Evaluate how well the proposed DSTSD achieves anomaly detection and
localization in Section \ref{subsubsec: anomaly_detection}.

\subsubsection{Spatio-temporal Mean Trend Prediction Accuracy \label{subsubsec: mean-trend}}

We compared the prediction accuracy of the learned spatio-temporal
metamodel in terms of the spatio-temporal mean trend prediction, i.e.,
$\mu(s,t)$. relative Mean square error (rMSE) of the predicted mean
for the prediction horizon $\Delta t$ ms were computed as 
\begin{equation}
\mathrm{rMSE}(\Delta t)=\frac{1}{NTn_{x}}\sum_{i,t}\frac{||\hat{\boldsymbol{\mu}}_{i,{t_{0}}}\left(t_{0}+\Delta t\right)-\boldsymbol{\mu}_{{i,t_{0}}}\left(t_{0}+\Delta t\right)||^{2}}{||\boldsymbol{\mu}_{{i}}\left(t_{0}\right)||^{2}},\label{eq: MSE}
\end{equation}
where $T$ is the length of the sequence in the testing data, $N$
is the number of testing samples, and $n_{x}$ is the length of the
spatial dimensions. For the benchmark method, we will compare with
the ST-SSD method \cite{yan2018real} in terms of future prediction
accuracy.

The prediction accuracy of the proposed DSTSD methods based on Conv-WaveNet
and Conv-LSTM architectures are calculated and compared using testing
data from both cases. For case 1, stimulation at every 800ms
is given to the 0th cell of the cable, which generates a series of
electrical waves propagating to both ends of the cable. For case 2,
a more complicated scenario is considered, where two stimulations
are given at the 1$^{\mathrm{th}}$ cell and the 600$^{\mathrm{th}}$
cell every 300ms. Two electrical waves are generated and propagate
toward each other and to both ends of the cable. For case 2, given
the 2nd stimulation is still inside the cardiac refractory period,
it will generate the wave. We will like to evaluate how the proposed
model is able to predict cardiac refractoriness behavior. For
the benchmark method, we will choose a time-series analysis but treat
each spatial dimension independently. The result of the rMSEs of $\Delta t$=
10ms, 100ms, and 200ms are shown in Table~\ref{Table: metamodel}.

\begin{table}[h!]
\centering \caption{Long-term mean trend prediction accuracy $\mathrm{MSE}$
for different prediction horizon $\Delta t=10,100,200$. }
\begin{tabular}{|c|c|c|c|}
\hline 
\multicolumn{4}{|c|}{Case 1}\tabularnewline
\hline 
Method  & $\Delta t=10$ & $\Delta t=100$ & $\Delta t=200$ \tabularnewline
\hline 
Conv-WaveNet  & \textbf{2.5e-5 (1e-5)}  & \textbf{9.9e-4 (3e-4)}  & \textbf{2.0e-3 (8e-4)}\tabularnewline
\hline 
Conv-LSTM  & 1.2e-3 (1e-4)  & 5e-3 (1e-3)  & 6e-3 (8e-4)\tabularnewline
\hline 
AutoRegressvie  & 1.6e-2 (3e-4)  & 1.9e-1 (6e-3)  & 4.2e-1 (1e-2)\tabularnewline
\hline 
\multicolumn{4}{|c|}{Case 2}\tabularnewline
\hline 
Method  & $\Delta t=10$ & $\Delta t=50$ & $\Delta t=100$\tabularnewline
\hline 
Conv-WaveNet  & \textbf{8.5e-3 (2e-2)}  & 8.5e-2 (8e-3)  & 2.0e-1 (2e-1)\tabularnewline
\hline 
Conv-LSTM  & 1.2e-2 \textbf{(}1.8e-2\textbf{)}  & \textbf{1.3e-2 (8e-3)}  & \textbf{4.9e-2 (6e-2)}\tabularnewline
\hline 
AutoRegressvie  & 2.7e-2 \textbf{(}2e-2\textbf{)}  & 3.1e-1 \textbf{(}1e-1\textbf{)}  & 6.3e-1 \textbf{(}3e-1\textbf{)}\tabularnewline
\hline 
\end{tabular}\label{Table: metamodel} 
\end{table}

\begin{figure}
\centering \subfloat[Simulation Model]{\includegraphics[width=0.33\linewidth]{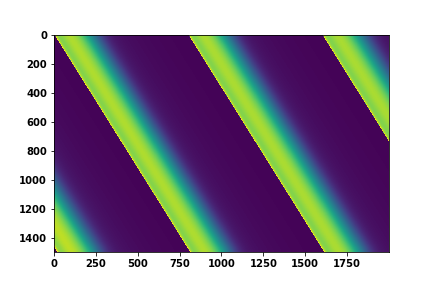}\label{case_I_800_true}

}\subfloat[Conv-LSTM ]{\includegraphics[width=0.33\linewidth]{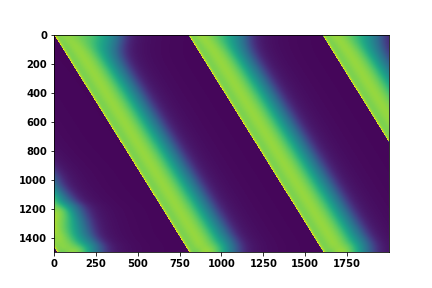}\label{case_I_800_lstm}

}\subfloat[Conv-WaveNet]{\includegraphics[width=0.33\linewidth]{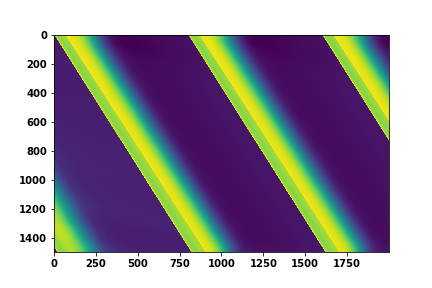}\label{case_I_800_cnn}

}

\subfloat[Simulation Model]{\includegraphics[width=0.33\linewidth]{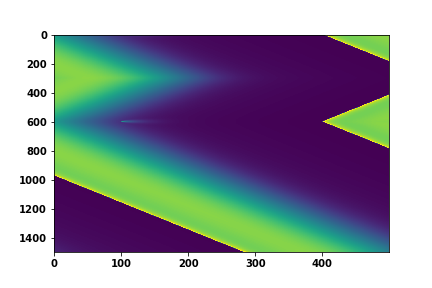}\label{case_II_300_true}

}\subfloat[Conv-LSTM]{\includegraphics[width=0.33\linewidth]{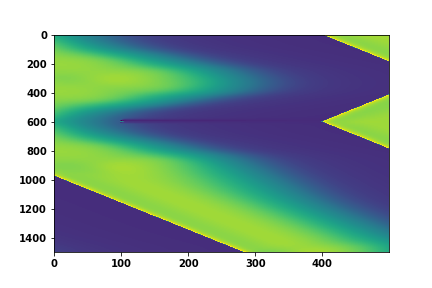}\label{case_II_300_LSTM}

}\subfloat[Conv-WaveNet]{\includegraphics[width=0.33\linewidth]{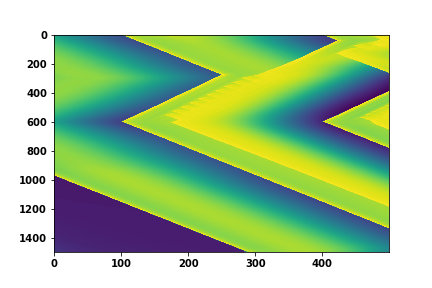}\label{case_II_300_cnn}

}

\caption{A comparison of Metamodel performance of the proposed
Conv-WaveNet and Conv-LSTM. Fig.~(a)-(c) illustrated case 1 and
(d)-(f) shows case 2. The results show that Conv-WaveNet does not
capture the cell refractoriness effect in case 2.}
\label{Fig: metamodel} 
\end{figure}

From Table~\ref{Table: metamodel}, we can conclude that in relatively
simple case (i.e., Case 1), Conv-WaveNet out-performs Conv-LSTM. However,
for a more complicated case (i.e., Case 2), where two waves merge,
and a new excitation is generated within the refractory period, Conv-LSTM
is able to predict this refractoriness quite accurately (i.e., a new
stimulation is not able to produce any waves.), but Conv-WaveNet failed
to predict this effect. For comparison, we also compare with the autoregressive
time-series model, which failed to capture any of the trends and result
in the largest error. To show this more clearly, we
also show the long-term prediction accuracy of both Conv-WaveNet and
Conv-LSTM in Fig.~\ref{Fig: metamodel}, which shows the prediction
of both Conv-WaveNet and Conv-LSTM compared to the true simulation
model. In the supplementary material, we also show the snapshots
and the video to demonstrate the performance of the proposed Conv-WaveNet
and Conv-LSTM in various scenarios. Overall, we can conclude that
Conv-LSTM works better and more robustly given different scenarios
compared to Conv-WaveNet, which only works well for relatively simple
cases.

Here, we would also like to report the computational time of the proposed
metamodels (i.e., Conv-WaveNet and Conv-LSTM) compared to the CRN
simulation model to simulate $1s$ of the cardiac signal. From Table~\ref{Table: time}, we can conclude that the metamodel is much faster
compared to the simulation model (i.e., 0.004s for Conv-LSTM compared
to 1.58s for the simulation model) without losing too much accuracy.
Conv-LSTM is faster than Conv-WaveNet due to its ability to use the
hidden variables to compress the historical observations.

\begin{table}[h!]
\centering \caption{The
computational time for the original Finite Element Simulation, the
proposed Conv-LSMT, and the proposed Conv-WaveNet to simulate $1s$
of the cardiac signal. }
\begin{tabular}{|c|c|c|c|}
\hline 
Method  & FEM Simulation  & Conv-LSTM  & Conv-WaveNet\tabularnewline
\hline 
Time  & 1.58s  & 0.004s  & 0.01s\tabularnewline
\hline 
\end{tabular}\label{Table: time} 
\end{table}

\subsubsection{Anomaly Detection Accuracy \label{subsubsec: anomaly_detection}}

In this subsection, we will compare the performance of our proposed
DSTSD with both architectures from Conv-WaveNet and Conv-LSTM (denoted
as 'DSTSD-ConvWaveNet' and 'DSTSD-ConvLSTM') with a few benchmark
methods in the literature in terms of anomaly detection. First, we
would like to compare with methods that only rely on residual of deep spatio-temporal
learning methods. We will use the deep learning architecture with
exactly the same architecture, namely the Conv-WaveNet and Conv-LSTM.
We also compare with the Hotelling $T^{2}$ method as the baseline
methods. For Hotelling $T^{2}$, it doesn't have the ability to model
the complicated spatio-temporal mean trend, so we use a simple difference
along the time dimension to remove the dynamic mean-trend beforehand.
We also try the moving average approach and do not find any improvement.
Finally, we compared the proposed methods with the ST-SSD methods,
which is another spatio-temporal decomposition method but with a fixed
smooth spatio-temporal basis \cite{yan2018real}.

For evaluation, we will compare the performance of the proposed DSTSD
with benchmark methods mentioned above in terms of average detection
delay and the localization accuracy. (i) To evaluate the detection
delay, we will use the out-of-control Average run length $ARL_{1}$,
which is defined as the average detection delay after the change occurs
with the fixed in-control $ARL_{0}$ as $0$. To evaluate the localization
accuracy, we will use three additional criteria after a shift is detected:
(ii) precision, defined as the proportion of detected anomalies that
are true anomalies; (ii) recall, defined as the proportion of the
anomalies that are correctly identified; (iii) F1-score, a single
criterion that combines the precision and recall \cite{van1979information}.
It is worth noting that only decomposition-based methods have the
ability to isolate the anomaly signals. Therefore, for non-decomposition-based
methods such as T2, Conv-LSTM, and Conv-WaveNet, we select a threshold
based on Otsu's method on the residual for source identification \cite{otsu1975threshold}.
Finally, the average values of these criteria and their standard deviation
over 100 simulation replications for $\delta=0.2$ and $\delta=0.3$
are given in Table~\ref{result2}.

\begin{table}[h]
\centering \caption{The anomaly detection means
and standard deviations (including precision, recall, F-1 score, and
ARL) for different change magnitudes $\delta=0.2$ and $\delta=0.3$
for different anomaly detection methods. }
\resizebox{\columnwidth}{!}{%
\begin{tabular}{|c|c|c|c|c|}
\hline 
\multicolumn{5}{|c|}{$\delta=0.3$}\tabularnewline
\hline 
Method  & precision  & recall  & F1-score & ARL \tabularnewline
\hline 
T2  & 0.188(0.087)  & 0.104(0.049)  & 0.131(0.061)  & 53.67(3.08) \tabularnewline
\hline 
Conv-LSTM  & 0.758(0.041)  & 0.677(0.034)  & 0.711(0.033)  & 24.16(2.67) \tabularnewline
\hline 
Conv-WaveNet  & 0.354(0.061)  & 0.250(0.046)  & 0.292(0.051)  & 45.35(4.62) \tabularnewline
\hline 
DSTSD-Conv-WaveNet  & 0.812(0.087)  & 0.625(0.077)  & 0.700(0.079)  & 29.75(4.34) \tabularnewline
\hline 
\textbf{DSTSD-ConvLSTM}  & \textbf{1.000(0.000)}  & \textbf{1.000(0.000)}  & \textbf{1.000(0.000)}  & \textbf{3.20(0.03)} \tabularnewline
\hline 
ST-SSD  & 0.06(0.007)  & 0.5(0.007)  & 0.1(0.012)  & 50.17(5.25) \tabularnewline
\hline 
\multicolumn{5}{|c|}{$\delta=0.2$}\tabularnewline
\hline 
T2  & 0.031(0.030)  & 0.021(0.020)  & 0.025(0.024)  & 49.21(1.59) \tabularnewline
\hline 
Conv-LSTM  & 0.448(0.087)  & 0.333(0.064)  & 0.373(0.071)  & 46.97(5.99) \tabularnewline
\hline 
Conv-WaveNet  & 0.250(0.625)  & 0.167(0.041)  & 0.200(0.050)  & 53.58(4.14) \tabularnewline
\hline 
DSTSD-ConvWaveNet  & 0.417(0.075)  & 0.292(0.050)  & 0.342(0.059)  & 42.06(4.47) \tabularnewline
\hline 
\textbf{DSTSD-ConvLSTM}  & \textbf{0.521(0.052)}  & \textbf{0.469(0.057)}  & \textbf{0.490(0.054)}  & \textbf{27.05(4.39)} \tabularnewline
\hline 
ST-SSD  & 0.008(0.001)  & 0.16(0.025)  & 0.10(0.012)  & 50.17(5.25) \tabularnewline
\hline 
\end{tabular}\label{result2}} 
\end{table}

From Table \ref{result2}, we can conclude that the proposed DSTSD-ConvLSTM
method achieves the best performance with the smallest detection delay
$ARL_{1}$. For example, when $\delta=0.3$, the $ARL_{1}=0.2$ for
the proposed DSTSD-ConvLSTM and the second-best Conv-LSTM has $ARL_{1}=24.16$.
Similarly, the proposed DSTSD-ConvLSTM has also the best performance
for localizing the source. For example, the $F=0.521$ for the proposed
DSTSD-ConvLSTM $\delta=0.2$ and the second-best Conv-LSTM has $F=0.448$.
In general, the proposed DSTSD methods are better than the prediction- based model considering both F1 and ARL. T2 doesn't work well due
to its inability to capture the complex spatial-temporal dynamics.
ST-SSD failed to detect the anomalies due to its strong smoothness
assumption, which is violated by the data generated from the CRN models.

The advantage of the performance is due to the following two reasons:
1) The ability to accurately capture the complex spatio-temporal patterns
of the mean trend. The importance of capturing spatio-temporal patterns
is demonstrated by comparing the Conv-LSTM, Conv-WaveNet, and the
ST-SSD. Conv-LSTM uses the combination of RNN and CNN, which gives
the best overall estimation of the spatio-temporal mean trend, which
is also shown in Table \ref{Table: metamodel}. Conv-WaveNet considers
complicated spatial structures. However, due to its use of the auto-regressive
model for the temporal structure, it works not as well as Conv-LSTM.
ST-SSD relies on the smoothness assumption with a fixed basis in both
the spatial dimension and the temporal dimension, which limits its
ability to capture and predict complex spatio-temporal dynamics. 2)
The ability to separate the anomaly signals considering the sparse
structure in the proposed DSTSD framework. In the proposed DSTSD methods
(i.e., both DSTSD-ConvLSTM and DSTSD-ConvWaveNet), we solve the inverse
problem using the buffered window approach, which achieves a better
estimation of the anomaly and leads to smaller $ARL_{1}$.

\begin{figure}[h!]
\centering \subfloat[Comparison of $ARL_{1}$]{\includegraphics[width=0.5\linewidth]{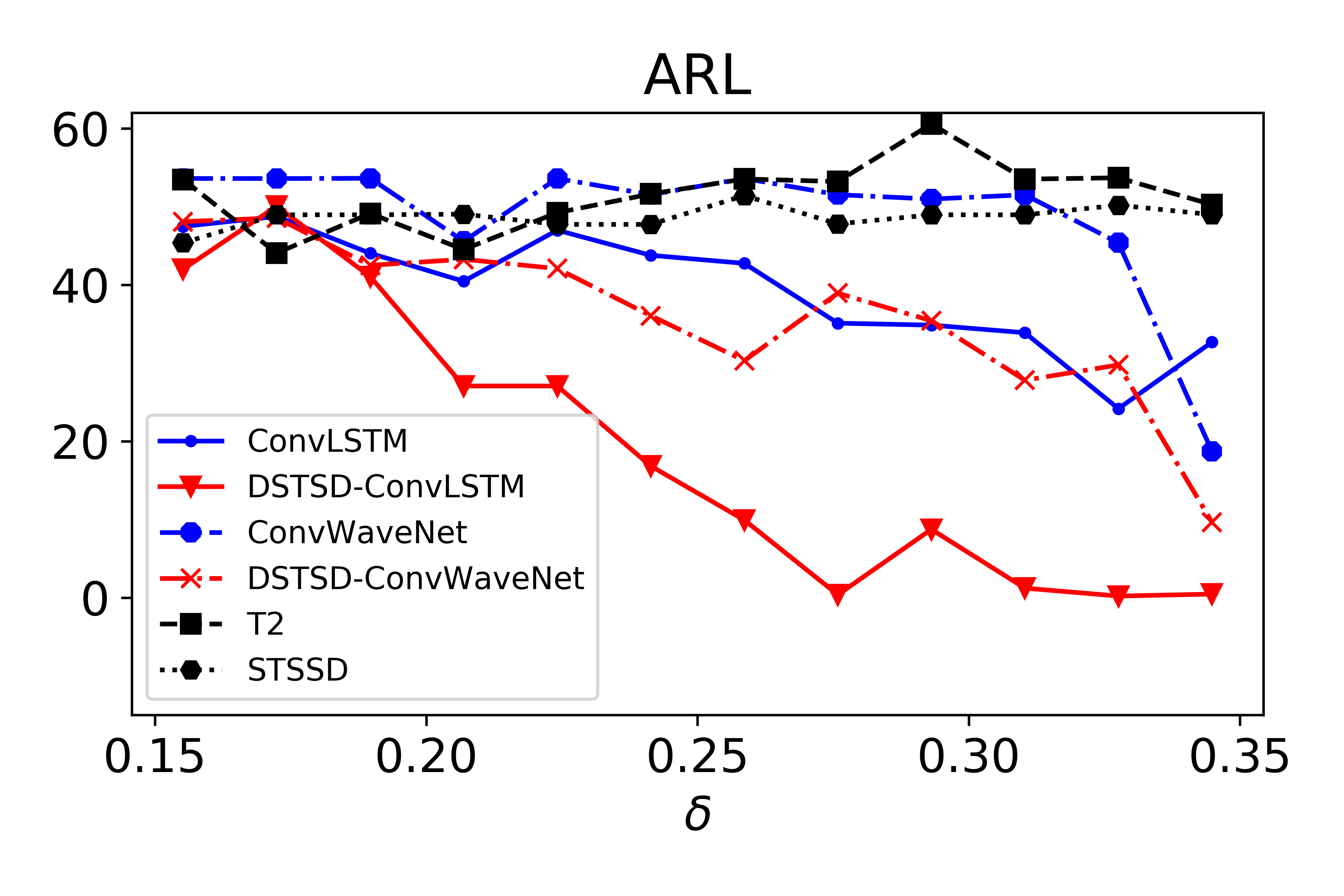}

}\subfloat[Comparison of F-1 score]{\includegraphics[width=0.5\linewidth]{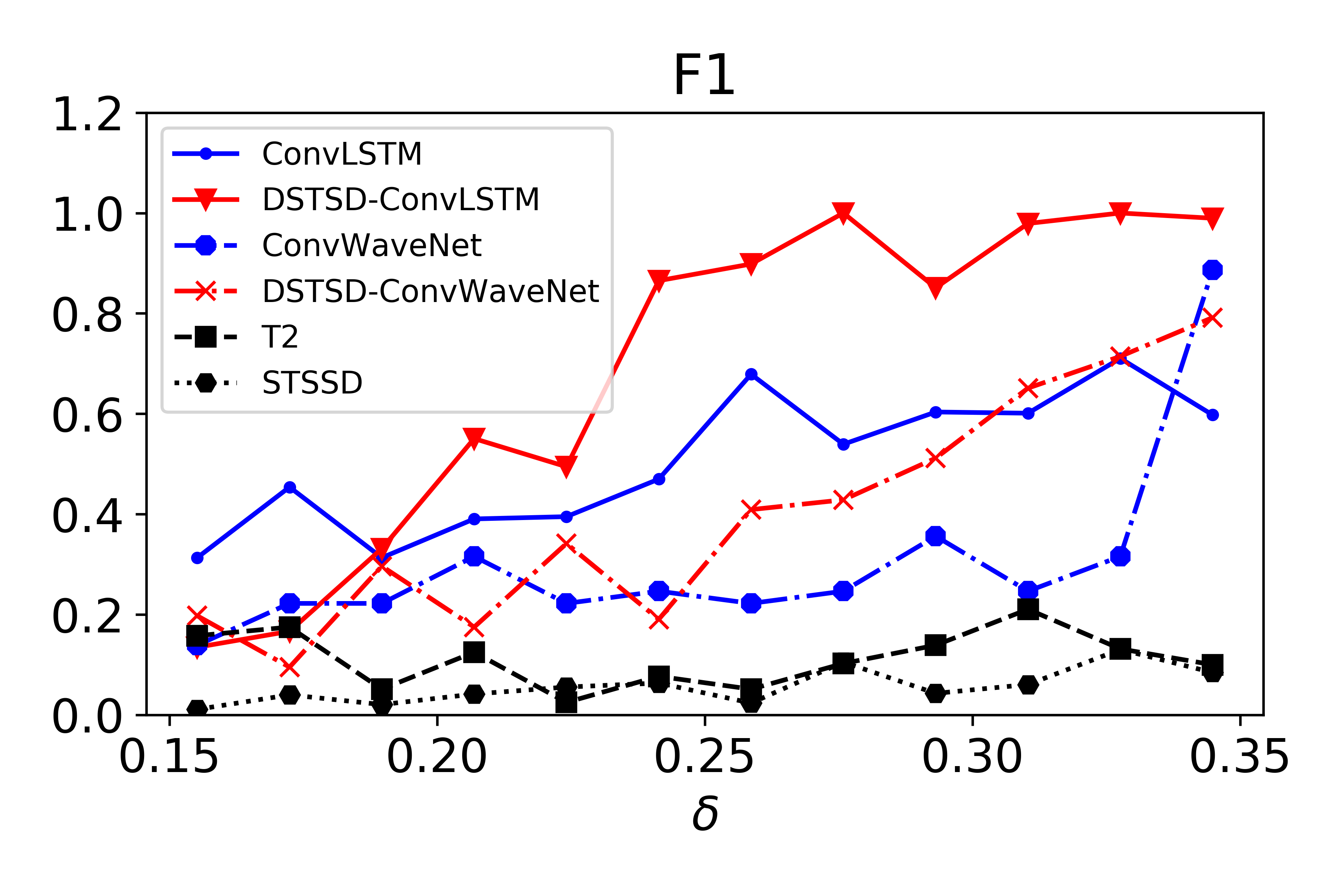}

}\caption{Evaluation criterion for proposed methods. Fig.~(a) and (b) shows the Out-of-control Average Run Length (ARL) and
the F-1 score of all methods according to different change magnitude
$\delta$. }
\label{result_arl} 
\end{figure}

We also perform a sensitivity analysis by comparing the proposed algorithm
with the benchmark on different change magnitudes $\delta$. Here,
we have generated abnormal cases with stimulation amplitude ranging
from 4.5 to 11 in 0.5 increments, which corresponds to the relative
magnitude $\delta$ from 0.15 to 0.35 in $0.015$ increase given the
noise level is $\sigma=1.5$. For each case, we generate 16 replications
for each delta. We further evaluate the $ARL_{1}$ and F1-score for
different benchmark methods under these magnitudes (i.e., different $\delta$). It is
clear that through solving the inverse problem, two DSTSD-based approaches
(i.e., Shown in red) get much better results than their corresponding
residual-based methods (i.e., shown in blue). Fig. \ref{result_arl}
shows that DSTSD-ConvLSTM can detect the change right away at $\delta=0.275$,
while the other methods got a large $ARL_{1}$ for every set up we
designed. From the result of the F1-score, we can see that DSTSD-ConvLSTM
can identify almost all anomalies as $\delta$ is close to 0.35 (i.e.,
F1-score is close to 1). 





\begin{figure}
\centering \subfloat[DSTSD-ConvLSTM]{\includegraphics[width=0.5\linewidth]{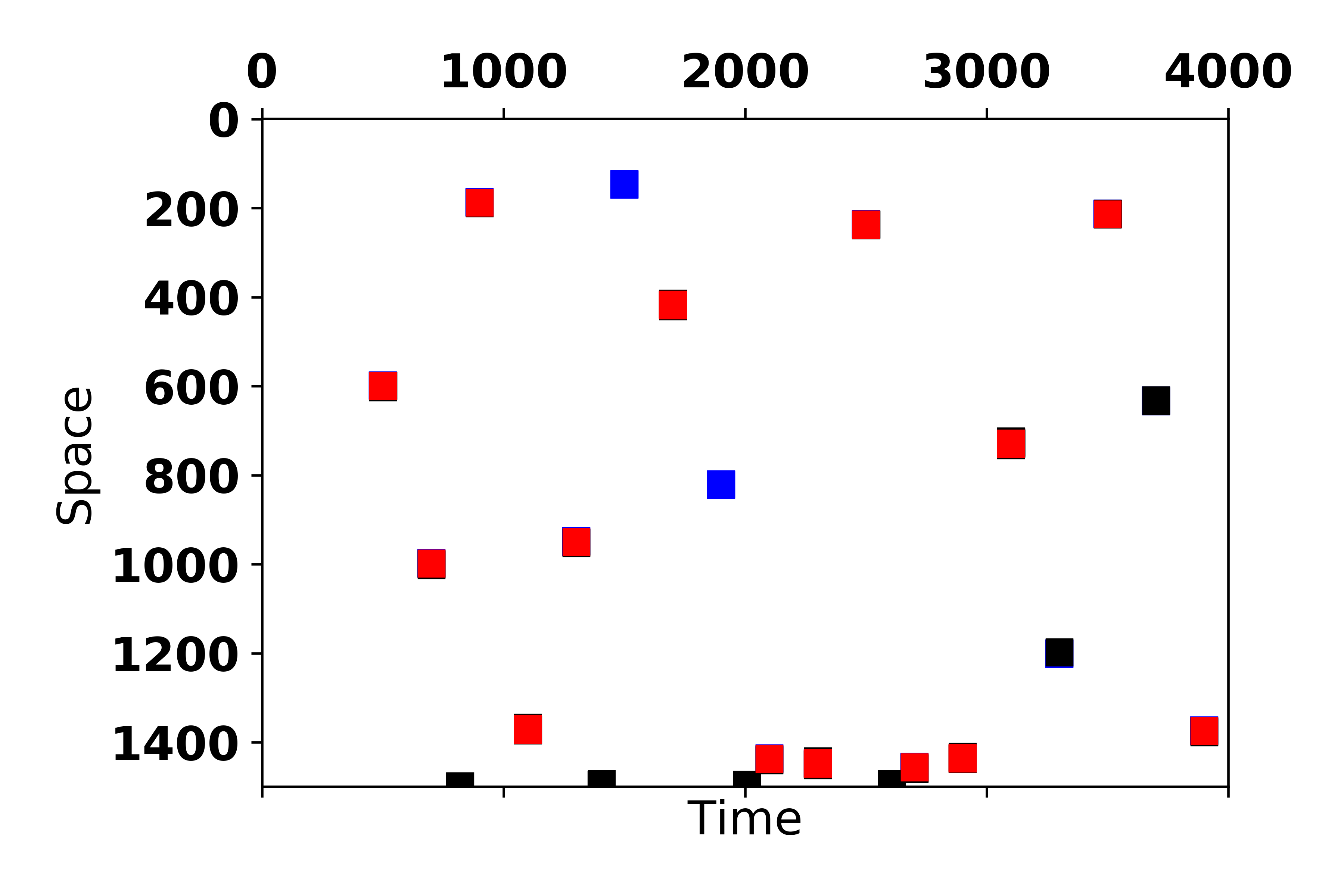}\label{case_I_result_lstm}

}\subfloat[DSTSD-ConvWaveNet ]{\includegraphics[width=0.5\linewidth]{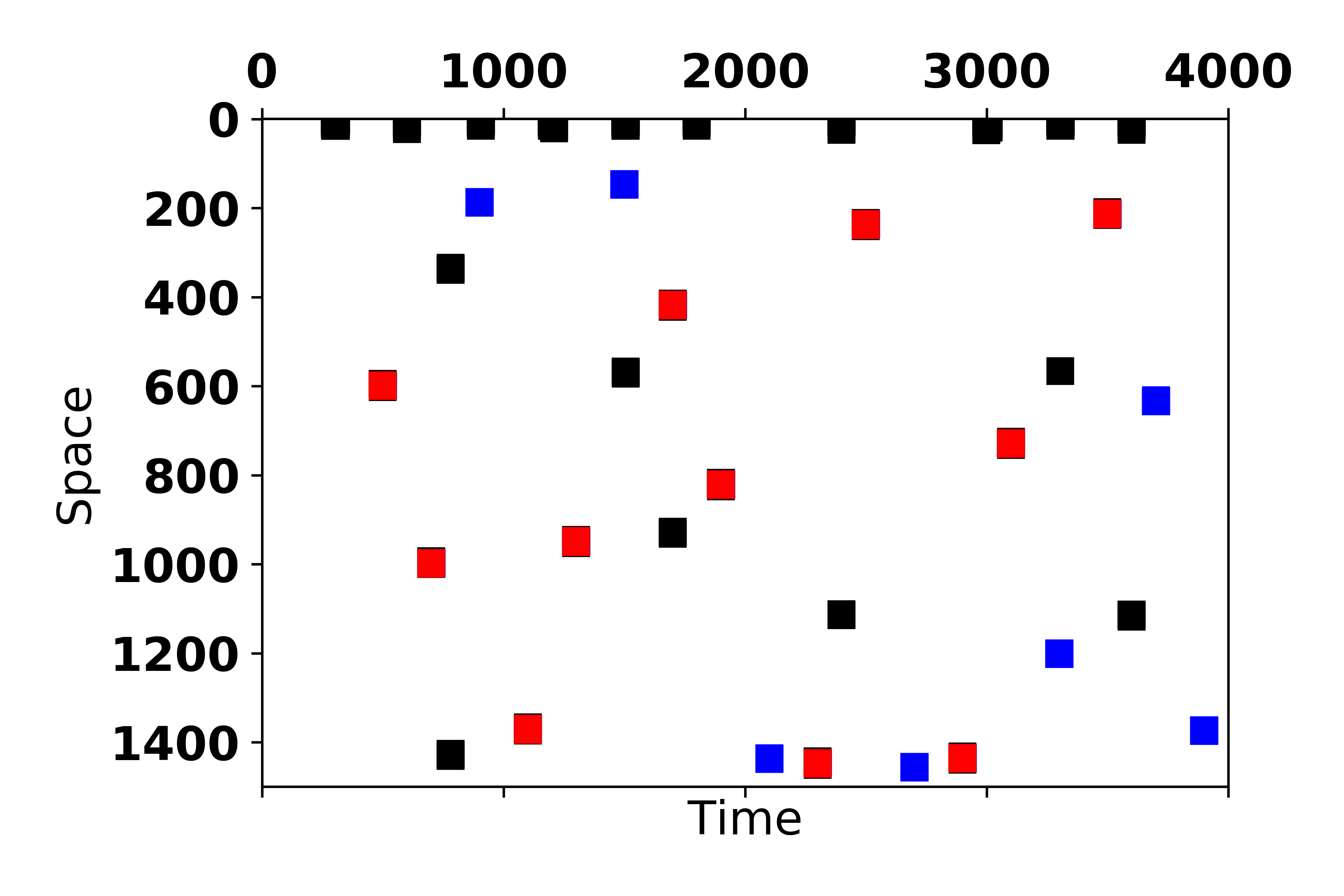}\label{case_I_lstm_sparse}

}\caption{Fig.~(a) and (b) show the comparison of the estimated
and actual stimulation for the proposed DSTSD-ConvLSTM and DSTSD-ConvWaveNet,
respectively. Here, the black square shows the false positives and
the blue square shows the false negative. The red square shows the
correct predictions. }
\label{predicted_anomaly} 
\end{figure}

To show how the proposed algorithm is able to isolate the source location.
An example of the detected source for Case I can be seen from Fig.
\ref{predicted_anomaly}. The correctly predicted stimulation points
(i.e., true positive) are shown in the red markers, missing stimulation
points (i.e., false negative) are shown in blue points, and incorrect
predictions (i.e., false positive) are shown in black markers. Through
solving the inverse problem, we are able to identify when and
where a stimulation happens. For Case I, there is a periodic stimulation
at a single location which is cell 1 (can be seen in Fig. \ref{case_I_Stimulation}
and Fig. \ref{case_I_simulation}). From the results, we can see that
both methods can identify almost all actual stimulations. In general,
DSTSD-ConvLSTM gives a better source localization result with fewer
false positives and false negatives compared to DSTSD-ConvWaveNet. In
summary, Conv-LSTM performs better than Conv-WaveNet.

\section{Conclusion \label{sec: conclusion}}

Identifying the cardiac cells that produce electrical impulses in
the system governed by complex spatio-temporal dynamics is an important
task. In this work, we first proposed a deep spatio-temporal sparse
decomposition approach to effectively decompose the original data
into a spatio-temporal mean trend as well as the sparse anomaly. To
effectively solve the optimization problem, the proximal gradient
descent algorithm is applied. To estimate the time and location of
the anomaly more accurately, we propose to solve the inverse problem
in a window-buffer approach to estimate the anomaly sources accurately.
Finally, a sequential likelihood ratio test was proposed to detect
the anomaly online. The proposed method is then validated through
the data set generated by the CRN model, which is widely used to simulate
the changes of transmembrane potential in human atrial cells. Through
extensive comparison, we showed that the proposed methods outperform
existing spatio-temporal modeling in terms of the spatio-temporal
mean trend prediction (i.e., metamodeling performance) , anomaly detection
and localization (i.e., anomaly detection performance). For future
works, we plan to combine the physical domain knowledge such as the
PDEs into the spatio-temporal model for a better generalization power. Furthermore, we are going to extend the current algorithm into multi-dimensional cases, which might be harder due to additional computational challenges.

\appendix

\section{Proof of Proposition 1 \label{apd: prop1}}
\begin{proof}
	Considering the loss function defined in (\ref{eq: lrobustloss}).
	\begin{align}
		l\left({\theta},{\theta_{a,i,t+1}}\right) & =\sum_{i,t}\parallel e_{i,t+1}-\theta_{a,i,t+1}\parallel^{2}\nonumber \\
		& +\lambda\mu_{t+1}'R\mu_{t+1}+\gamma\|\theta_{a,i,t+1}\|_{1},\label{eq: lrobustloss}
	\end{align}
	where the residual can be defined in (\ref{eq: residualr}). 
	\begin{equation}
		e_{i,t+1}=y_{i,t+1}-g\left(\{y_{i,t'}\}_{t'\leq t};\theta\right)-\mu_{i,t}-r_{i,t+1}.\label{eq: residualr}
	\end{equation}
	Here, the loss function in (\ref{eq: lrobustloss}) can be decoupled
	into each pair of $(i,t)$ individually. Therefore, each $\theta_{a,i,t+1}$
	can be solved individually by optimizing 
	\begin{equation}
		\hat{\theta}_{a,i,t+1}=\arg\min_{\theta_{a,i,t+1}}\parallel e_{i,t+1}-\theta_{a,i,t+1}\parallel^{2}+\gamma\|\theta_{a,i,t+1}\|_{1}\label{eq: softthresholdsqure}
	\end{equation}
	Finally, (\ref{eq: softthresholdsqure}) can be solved by 
	\begin{equation}
		\hat{\theta}_{a,i,t+1}=S_{\gamma/2}(e_{i,t+1}),\label{eq: softresidual}
	\end{equation}
	which is the same as the update step in Proposition 1 given the definition
	of $e_{i,t+1}$ in (\ref{eq: residualr}).
	
	Finally, we can plug in the solution of $\hat{\theta}_{a,i,t+1}$
	as in (\ref{eq: softresidual}) into $l\left(\theta,\theta_{a,i,t+1}\right)$
	defined in (\ref{eq: lrobustloss}). We will consider two different
	cases: 
\end{proof}
\begin{enumerate}
	\item If $|e_{i,t+1}|>\gamma/2$, $\hat{\theta}_{a,i,t+1}=(|e_{i,t+1}|-\gamma/2)\mathrm{sgn}(e_{i,t+1})$,
	therefore, the loss in (\ref{eq: softresidual}), which is related
	to $\theta_{a,i,t+1}$ is 
	\begin{align*}
		& \parallel e_{i,t+1}-\theta_{a,i,t+1}\parallel^{2}+\gamma\|\theta_{a,i,t+1}\|_{1}\\
		= & \frac{\gamma^{2}}{4}+\gamma(|e_{i,t+1}|-\gamma/2)\\
		= & \gamma|e_{i,t+1}|-\frac{\gamma^{2}}{4}.
	\end{align*}
	
	\begin{enumerate}
		\item If $|e_{i,t+1}|<\gamma/2$, $\hat{\theta}_{a,i,t+1}=0$, 
		\begin{align*}
			& \parallel e_{i,t+1}-\theta_{a,i,t+1}\parallel^{2}+\gamma\|\theta_{a,i,t+1}\|_{1}\\
			= & \|e_{i,t+1}\|^{2}.
		\end{align*}
		In conclusion, this implies: 
		\begin{equation}
			\parallel e_{i,t+1}-\theta_{a,i,t+1}\parallel^{2}+\gamma\|\theta_{a,i,t+1}\|_{1}=\rho(e_{i,t+1}),\label{eq: huberloss_plug}
		\end{equation}
		where $\rho(x)=\begin{cases}
			x^{2} & |x|\leq\frac{\gamma}{2}\\
			\gamma|x|-\frac{\gamma^{2}}{4} & |x|>\frac{\gamma}{2}
		\end{cases}$ is the Huber loss. 
	\end{enumerate}
\end{enumerate}
\begin{proof}
	By plugging in the definition of (\ref{eq: huberloss_plug}) into
	(\ref{eq: lrobustloss}), we have $l_{r}(\theta)=\sum_{i}\sum_{t}\left(\rho(e_{i,t+1})+\lambda\mu_{t}'R\mu_{t}\right)$.
	By plugging in the definition of $e_{i,t+1}$, we can prove proposition
	1. 
\end{proof}

\section{Proof of Proposition 2 \label{apd: prop2}}
\begin{proof}
	Considering the loss function in 
	\[
	\min_{\theta_{a,t},t\in[T,T+w]}l_{T\rightarrow T+w}(\{\theta_{a,t}\})+\gamma\sum_{t=T}^{T+w}\|\theta_{a,t}\|_{1}
	\]
	Since most neural network architecture is Lipschitz continuous \cite{virmaux2018lipschitz},
	$l_{T\rightarrow T+w}(\{\theta_{a,t}\})$ is Lipschitz continuous.
	We assume the Lipschitz constance is $L$. Therefore, according to
	the proximal gradient procedure, for each $\theta_{a,t}$ at each
	iteration $k$, we can minimize the upper bound of $l_{T\rightarrow T+w}(\{\theta_{a,t}\})+\gamma\sum_{t=T}^{T+w}\|\theta_{a,t}\|_{1}$
	at iteration $k$ as 
	\begin{align}
		\theta_{a,t}^{(k)} & =\arg\min_{\theta_{a,t}}l_{T\rightarrow T+w}(\theta_{a,t}^{(k-1)})\nonumber \\
		& +\left\langle \theta_{a,t}-\theta_{a,t}^{(k-1)},\frac{\partial l_{T\rightarrow T+w}(\{\theta_{a,t}\})}{\partial\theta_{a,t}}\right\rangle \nonumber \\
		& +\frac{L}{2}\|\theta_{a,t}-\theta_{a,t}^{(k-1)}\|^{2}+\gamma\|\theta_{a,t}\|_{1}\label{eq: Lipchitz}
	\end{align}
	This can be solved in closed-form as 
	\[
	\theta_{a,t}=S_{\gamma/2}(\theta_{a,t}^{(k-1)}-\frac{1}{L}\frac{\partial l_{T\rightarrow T+w}(\{\theta_{a,t}\})}{\partial\theta_{a,t}}).
	\]
	This is the same as the equation in Proposition 2 as the step size
	$c=1/L$. 
\end{proof}

\section{Conv-LSTM Architecture \label{apd: convlstmarchitecture}}

The architectures of Conv-LSTM is shown as follows. 

\begin{table}[H]
	\caption{Conv-LSTM Architecture}
	\centering %
	\begin{tabular}{|c|c|c|}
		\hline 
		Layer  & Conv Kernel Size  & Out-channel\tabularnewline
		\hline 
		Conv1DLSTM+Padding(2)  & $15$  & 10\tabularnewline
		\hline 
		Conv1D + ReLU + Padding(2)  & $3$  & 5\tabularnewline
		\hline 
		Conv1D  & $3$  & 1\tabularnewline
		\hline 
	\end{tabular}
\end{table}

Here, the notation is introduced as follows:  
\begin{itemize}
	\item Conv1DLSTM to refer to the LSTM with 1-D convolution as defined \cite{RN10}.
	
	\item Padding(2): implies the Replication Padding with size $2$ 
	\item ReLU: Rectified Linear Unit defined as $\mathrm{relu}(x)=\max(x,0)$. 
\end{itemize}
Considering the efficiency of computation, SGD is used as an optimizer
for initial steps. During the SGD training process, we select the
learning rate as 0.001 and momentum as 0.9. We found after gone through
the entire dataset for 10 epoches, the algorithm becomes very unstable.
Thus, we switch to AdamW when the fluctuation happens. AdamW is an
optimizer improved based on Adam optimizer which have much better
generalization behavior. Furthermore, for training the Conv-LSTM,
the gradient will be clipped with max norm 0.01 during each iteration
avoiding the gradient explosion issue. 

\section{Conv-Wavenet Architecture \label{apd: convwavenetarchitecture}}

The architectures of Conv-Wavenet is shown as follows. 

\begin{table}[H]
	\caption{Conv-Wavenet Architecture}
	\centering %
	\begin{tabular}{|c|c|c|}
		\hline 
		Layers  & Kernel Size  & Dilation size\tabularnewline
		\hline 
		DilatedConv2D + PReLU+Padding(0,2)  & $(2,17)$  & $(1,1)$\tabularnewline
		\hline 
		DilatedConv2D + PReLU+Padding(0,2)  & $(2,17)$  & $(2,1)$\tabularnewline
		\hline 
		DilatedConv2D + PReLU+Padding(0,2)  & $(2,17)$  & $(4,1)$\tabularnewline
		\hline 
		DilatedConv2D + PReLU+Padding(0,2)  & $(2,17)$  & $(8,1)$\tabularnewline
		\hline 
		DilatedConv2D + PReLU+Padding(0,2)  & $(2,17)$  & $(16,1)$\tabularnewline
		\hline 
		DilatedConv2D + PReLU+Padding(0,2)  & $(2,17)$  & $(32,1)$\tabularnewline
		\hline 
		DilatedConv2D + PReLU+Padding(0,2)  & $(2,17)$  & $(64,1)$\tabularnewline
		\hline 
	\end{tabular}
\end{table}

\begin{figure}
	\includegraphics[width=1\linewidth]{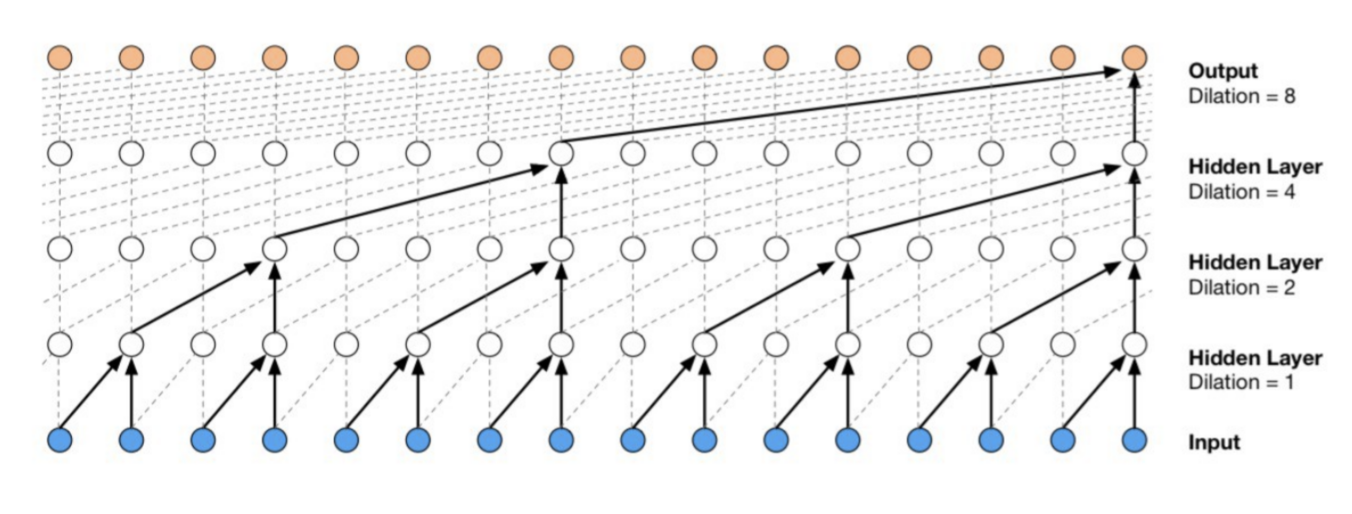}
	
	\caption{A Demo of WaveNet Architecture \cite{oord2016wavenet}}
	
	\label{Fig: WaveNet} 
\end{figure}

Here, the notation is introduced as follows:  
\begin{itemize}
	\item DilatedConv2D: to refer to the 2D dilated convolution and the kernel
	size is in the format of (Time kernel size, Space kernel size). We
	will kept the kernel size as $(2,5)$ and increase the dilation kernel
	size as $(2^{d},1)$, where $d$ is the dept of the network.  
	\item Padding(0,2): We will use the replication padding as well with time
	padding size $0$ and kernel padding size $2$.  
	\item PReLU: Parametric Rectified Linear Unit defined as $\mathrm{prelu}(x)=\max(x,0)+a\min(x,0)$.
	where a is a learnable parameter that control the PReLU layer. When
	$a=0$, the layer is same as ReLU. When $a>0$, the layer becomes
	a leaky ReLU layer. 
\end{itemize}
This architecture is inspired by WaveNet \cite{oord2016wavenet},
which use exponential growth dilation to increase the reception field
so that it can model the long-term dependency of the time series data
as shown in Figure \ref{Fig: WaveNet}. Here, the major difference
with the WaveNet is that we uses the 2D convolution to replace the
1D convolution to model the complex spatial-temporal relationship. 

Here, we also apply both SGD and AdamW as the optimizer and switch
between two optimizers to ensure stable learning curve. We select
same learning rate as 0.001. Beyond that we also apply a Gaussian
noise to the original data to make the model more robust against the
noise for phase II analysis. From our experience, adding the noise
criterion will help the behavior of the model improve a lot. 

\section{Metalearning Performance and Video \label{apd:video}}

\begin{figure}
	\subfloat[Case 1 with 300ms cycle]{\includegraphics[width=0.5\linewidth]{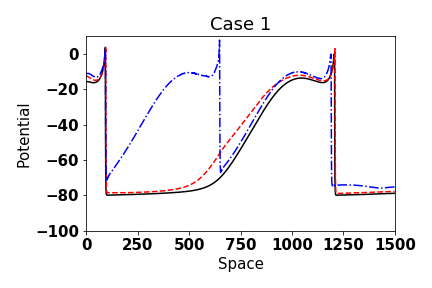}
		
	}\subfloat[Case 1 with 800ms cycle]{\includegraphics[width=0.5\linewidth]{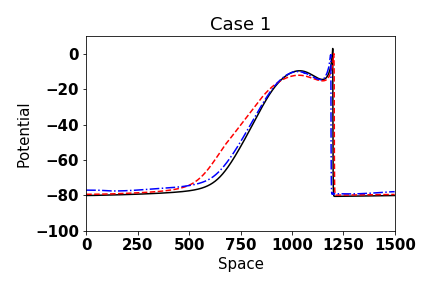}
		
	}
	
	\subfloat[Case 2 with 300ms cycle]{\includegraphics[width=0.5\linewidth]{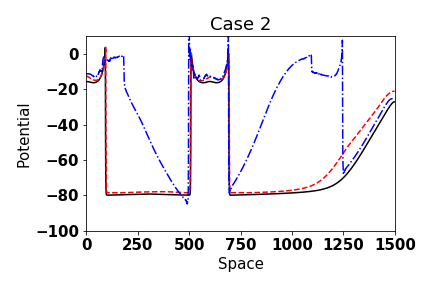}
		
	}\subfloat[Case 2 with 800ms cycle]{\includegraphics[width=0.5\linewidth]{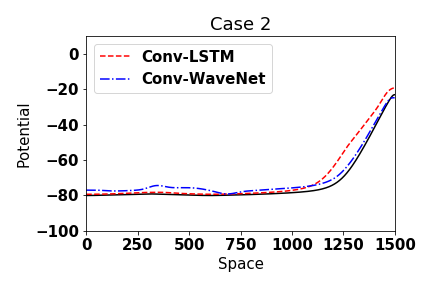}
		
	}
	
	\caption{100-time-ahead Prediction for Conv-LSTM and Conv-WaveNet, the full
		videos are uploaded in the Supplementary Material}
	\label{fig: snapshot} 
\end{figure}

We also uploaded the video in the supplementary material on the performance
of the metamodels with Conv-LSTM and Conv-WaveNet architectures. There
are four videos uploaded, which corresponding to Case 1 and Case 2
with 300ms and 800ms regular stimulation with 550ms-ahread prediction.
The snapshots of the videos are shown in Figure \ref{fig: snapshot}.
In Figure \ref{fig: snapshot} (a) and (c), the stimulation is still
within the cell refractory period and should not produce the wave.
Only Conv-LSTM is able to capture this trend, where Conv-WaveNet falsely
predict the wave is generated.

{} 
\section{Simulation resutls for FHN model}
\begin{figure}[h!]
	\centering \subfloat[Case 1 prediction results]{\includegraphics[width=0.35\linewidth]{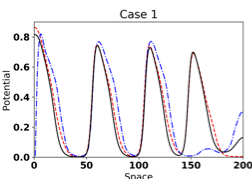} \label{case_II_simulation}
		
	}\subfloat[Case 2 prediction results]{\includegraphics[width=0.35\linewidth]{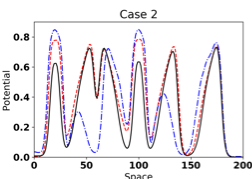}\label{case_II_anomaly}
		
	}
	\caption{Comparison of the wave pattern at $\delta t=200ms$ predicted by AR-CNN and Conv-LSTM vs true data}
	\label{Fig: 100_case1} 
\end{figure}

\begin{figure}[h!]
	\centering \subfloat[Case 1 prediction results]{\includegraphics[width=0.35\linewidth]{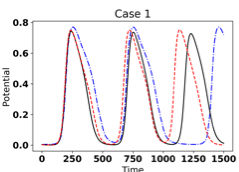} \label{case_II_Stimulation}
		
	}\subfloat[Case 2 prediction results]{\includegraphics[width=0.35\linewidth]{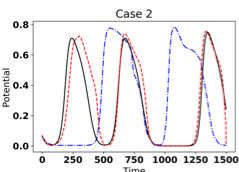}\label{case_II_2d}
		
	}\caption{Comparison of the 1500ms-time ahead prediction of AR-CNN, Conv-LSTM, and true data. The origin 0 is the start of prediction time $t_0$}
	\label{Fig: 1500_case1} 
\end{figure}

In this study, we have added another simulation case, where the spatio-temporal dynamics is simulated by Fitz-Hugh-Nagumo model (FHN). The following visualization is actually for the results of FHN model is given here. We also show more intermediate results about signal prediction as shown in Figure.\ref{Fig: 100_case1} and Figure.\ref{Fig: 1500_case1} in the following. Table.\ref{result_old} shows the prediction accuracy of the proposed model.

\begin{table}[h!]
	\centering \caption{Long-Term prediction accuracy in the noiseless case}
	\begin{tabular}{|c|l|l|l|}
		\hline
		\multicolumn{4}{|c|}{\textbf{Case I}}                                                                                                \\ \hline
		\textbf{$\Delta t(ms)$} & \multicolumn{1}{c|}{\textbf{AR-CNN}} & \multicolumn{1}{c|}{\textbf{Conv-LSTM}} & \multicolumn{1}{c|}{\textbf{AR}} \\ \hline
		5                & 2.8e-5 (1e-5)                        & 3.9e-5 (1e-5)                           & 2.7e-4 (7e-5)                    \\ \hline
		10               & 9.0e-5 (3e-5)                        & 4.1e-5   (1e-5)                         & 1.1e-3 (3e-4)                    \\ \hline
		50               & 1.9e-3 (1e-3)                        & 5.0e-4   (2e-4)                         & 0.023 (0.03)                     \\ \hline
		100              & 5e-3 (1e-3)                          & 1.0e-3   (6e-4)                         & 0.064 (0.02)                     \\ \hline
		200              & 1e-2 (4e-3)                          & 4e-3   (6.2e-4)                         & 0.140 (0.04)                     \\ \hline
		\multicolumn{4}{|c|}{\textbf{Case 2}}                                                                                                \\ \hline
		\textbf{delta t} & \multicolumn{1}{c|}{\textbf{AR-CNN}} & \multicolumn{1}{c|}{\textbf{Conv-LSTM}} & \multicolumn{1}{c|}{\textbf{AR}} \\ \hline
		5                & 3.8e-5   (2e-5)                      & 3.9e-5 (1e-5)                           & 2.7e-4 (7e-5)                    \\ \hline
		10               & 8.5e-5 (9e-5)                        & 4.0e-5 (1e-5)                           & 1.1e-3 (3e-4)                    \\ \hline
		50               & 1.6e-3 (2e-3)                        & 5.0e-4 (2e-3)                           & 0.02 (0.03)                      \\ \hline
		100              & 5.9e-3 (1e-3)                        & 1.0e-3   (7e-4)                         & 0.06 (0.02)                      \\ \hline
		200              & 0.013 (8e-3)                         & 1.5e-3   (6e-4)                         & 0.14 (0.04)                      \\ \hline
	\end{tabular}\label{result_old}
\end{table}

\section{Notation Table}

\global\long\def\nomname{Notations}%
\printnomenclature[3em]\nomenclature{$\mathbf{\mu}_t$}{Functional mean of the transmembrane potential}\nomenclature{$\mathbf{a}_t$}{Abnormal stimulation}\nomenclature{$\mathbf{r}_t$}{Regular stimulation}\nomenclature{$\mathbf{c}_t$}{Total stimulation}\nomenclature{$\mathbf{e}_t$}{Noise magnitude}\nomenclature{$\mathbf{y}_t$}{Observed potential}\nomenclature{$\boldsymbol{\theta}$}{Parameters for the estimation function $g$}\nomenclature{$\mathbf{B}_a$}{Anomaly basis}\nomenclature{$\boldsymbol{\theta}_{a,t}$}{Spatio-temporal coefficients of the anomaly at time t}\nomenclature{$\mathbf{R}$}{Regalarization matrix}\nomenclature{$\mathbf{z}_{f_t}, \mathbf{z}_{h_{t}}, \mathbf{z}_{i_t}, \mathbf{z}_{c_t},\mathbf{z}_{o_t}$}{Gate and cell state variables inside the LSTM model}\nomenclature{$\boldsymbol{\theta}_{W_f}, \boldsymbol{\theta}_{U_f}, \boldsymbol{\theta}_{V_f},\boldsymbol{\theta}_{W_i}, \boldsymbol{\theta}_{U_i}, \boldsymbol{\theta}_{V_i}, \boldsymbol{\theta}_{W_c}, \boldsymbol{\theta}_{U_c}, \boldsymbol{\theta}_{V_c},\boldsymbol{\theta}_{W_o}, \boldsymbol{\theta}_{U_o}, \boldsymbol{\theta}_{V_o}$}{parameters for conv-LSTM model}\nomenclature{$\mathbf{A}_{s,t}$}{Abnormal stimulation matrix}

\bibliographystyle{IEEEtran}
\bibliography{monitor}

\end{document}